%% file: main.tex
\documentclass{article}

\usepackage[final]{neurips_2023}

\usepackage[utf8]{inputenc} % allow utf-8 input
\usepackage[T1]{fontenc}    % use 8-bit T1 fonts
\usepackage{hyperref}       % hyperlinks
\usepackage{url}            % simple URL typesetting
\usepackage{booktabs}       % professional-quality tables
\usepackage{amsfonts}       % blackboard math symbols
\usepackage{nicefrac}       % compact symbols for 1/2, etc.
\usepackage{microtype}      % microtypography
\usepackage{xcolor}         % colors
\definecolor{ForestGreen}{RGB}{34, 139, 34}
\definecolor{Plum}{RGB}{221,160,221}
\definecolor{NavyBlue}{RGB}{0,0,128}

\usepackage{graphicx}
\usepackage{subfigure}

\usepackage{algpseudocode}
\usepackage{algorithm}
\usepackage[ruled,algo2e]{algorithm2e}
\usepackage{amsmath}
\usepackage{mathtools}
\usepackage{amsthm}
\usepackage{amssymb}
\usepackage{bm}
\usepackage{ulem}

\usepackage[capitalize,noabbrev]{cleveref}

\usepackage{anyfontsize}

\theoremstyle{plain}
\newtheorem{theorem}{Theorem}[section]
\newtheorem{proposition}[theorem]{Proposition}

\theoremstyle{definition}
\newtheorem{definition}[theorem]{Definition}
\newtheorem{assumption}[theorem]{Assumption}
\theoremstyle{remark}

\usepackage{enumitem}
\setlist[itemize]{itemsep = 1pt, topsep=0pt}
\setlist[enumerate]{itemsep= 1pt, topsep=0pt}

\title{FedGCN: Convergence-Communication Tradeoffs in Federated Training of Graph Convolutional Networks}

\author{%
  Yuhang Yao \\
  Carnegie Mellon University\\
  \texttt{yuhangya@andrew.cmu.edu} \\
  \And
  Weizhao Jin \\
  University of Southern California\\
  \texttt{weizhaoj@usc.edu} \\
  \And
  Srivatsan Ravi \\
  University of Southern California\\
  \texttt{sravi@isi.edu} \\
  \And
  Carlee Joe-Wong \\
  Carnegie Mellon University\\
  \texttt{cjoewong@andrew.cmu.edu} \\
}

\begin{document}

\maketitle

\input{0-abstract}

\input{1-introduction}
\input{2-related_work}
\input{3-problem_statement}

\input{4-fedgcn}

\input{5-theory}

\input{6-experiment}

\input{7-conclusion}

\input{8-use_case}

\input{9-future_directions_and_applications}

\section{Acknowledgement}
This research was supported by the National Science Foundation under grant CNS-1909306, Cloudbank support through CNS-1751075, and the Lee-Stanziale Ohana Fellowship by the ECE department at Carnegie Mellon University. The authors would like to thank Jiayu Chang, Cynthia (Xinyi) Fan, and Shoba Arunasalam for helping with the coding.

%\newpage
\bibliography{ref}
\bibliographystyle{ACM-Reference-Format}

%%%%%%%%%%%%%%%%%%%%%%%%%%%%%%%%%%%%%%%%%%%%%%%%%%%%%%%%%%%%%%%%%%%%%%%%%%%%%%%
%%%%%%%%%%%%%%%%%%%%%%%%%%%%%%%%%%%%%%%%%%%%%%%%%%%%%%%%%%%%%%%%%%%%%%%%%%%%%%%
% SUPPLEMENTAL CONTENT AS APPENDIX AFTER REFERENCES
%%%%%%%%%%%%%%%%%%%%%%%%%%%%%%%%%%%%%%%%%%%%%%%%%%%%%%%%%%%%%%%%%%%%%%%%%%%%%%%
%%%%%%%%%%%%%%%%%%%%%%%%%%%%%%%%%%%%%%%%%%%%%%%%%%%%%%%%%%%%%%%%%%%%%%%%%%%%%%%
\newpage
\input{appendix}
% \bibliography{ref}
% \bibliographystyle{ACM-Reference-Format}
%%%%%%%%%%%%%%%%%%%%%%%%%%%%%%%%%%%%%%%%%%%%%%%%%%%%%%%%%%%%%%%%%%%%%%%%%%%%%%%
%%%%%%%%%%%%%%%%%%%%%%%%%%%%%%%%%%%%%%%%%%%%%%%%%%%%%%%%%%%%%%%%%%%%%%%%%%%%%%%

\end{document}

%% file: 0-abstract.tex
\begin{abstract}
    %%%%% New abstract, old one commented out below
    Methods for training models on graphs distributed across multiple clients have recently grown in popularity, due to the size of these graphs as well as regulations on keeping data where it is generated. However, the cross-client edges naturally exist among clients. Thus, distributed methods for training a model on a single graph incur either significant communication overhead between clients or a loss of available information to the training. We introduce the Federated Graph Convolutional Network (FedGCN) algorithm, which uses federated learning to train GCN models for semi-supervised node classification with fast convergence and little communication. Compared to prior methods that require extra communication among clients at each training round, FedGCN clients only communicate with the central server in one pre-training step, greatly reducing communication costs and allowing the use of homomorphic encryption to further enhance privacy. We theoretically analyze the tradeoff between FedGCN's convergence rate and communication cost under different data distributions. 
    Experimental results show that our FedGCN algorithm achieves better model accuracy with 51.7\% faster convergence on average and at least 100$\times$ less communication compared to prior work\footnote{Code in \url{https://github.com/yh-yao/FedGCN}}.%, 51.7\% faster convergence on average, %\footnote{Code in \url{https://github.com/yh-yao/FedGCN}}
    %
    %Distributed methods for training models on graph datasets have recently grown in popularity, due to the size of graph datasets like social network structures. However, the graphical structure of this data means that it cannot be disjointly partitioned between different learning clients, leading to either significant communication overhead between clients or a loss of information available to the training method. We introduce Federated Graph Convolutional Network (FedGCN), an efficient framework for training GCN (graph convolutional networks) with optimized convergence rate and communication cost in Federated learning. Compared with distributed settings which require communication among clients at each iteration, FedGCN preserves the privacy of client data and only needs communication at the initial step, which greatly reduces communication cost and speeds up the convergence rate without information loss. We then theoretically analyze FedGCN's convergence rate and communication cost tradeoff of different orders of approximation under different distributions (i.e. IID, Partial-IID, Non-IID), showing that different distributions result in different tradeoff points. Our theoretical framework quantifying these tradeoffs can be generally used for the analysis of all edge-completion-based algorithms. Experimental results over both simulated and real-world datasets demonstrate the effectiveness of our algorithm and validate our theoretical analysis. 
    
    %\carlee{what does approximation rates mean?}
    %orders of approximation
    %1-order approximation, 0-order approximation
\end{abstract}

%% file: 1-introduction.tex
\section{Introduction}

\begin{figure}[t]
    \centering
    \includegraphics[width=0.65\textwidth]{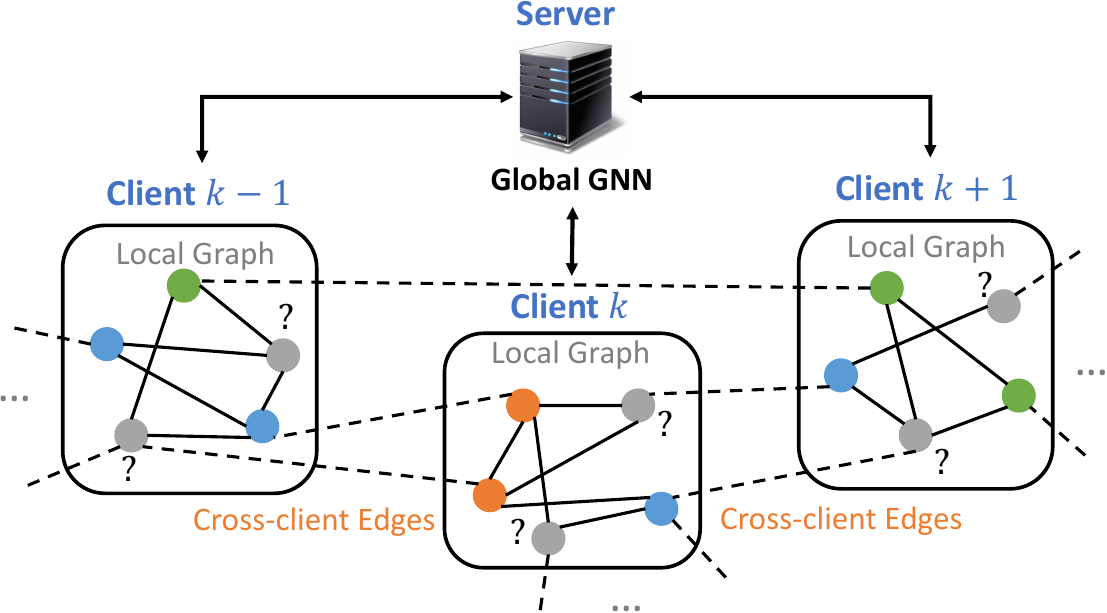}
    \includegraphics[width=0.27\textwidth]{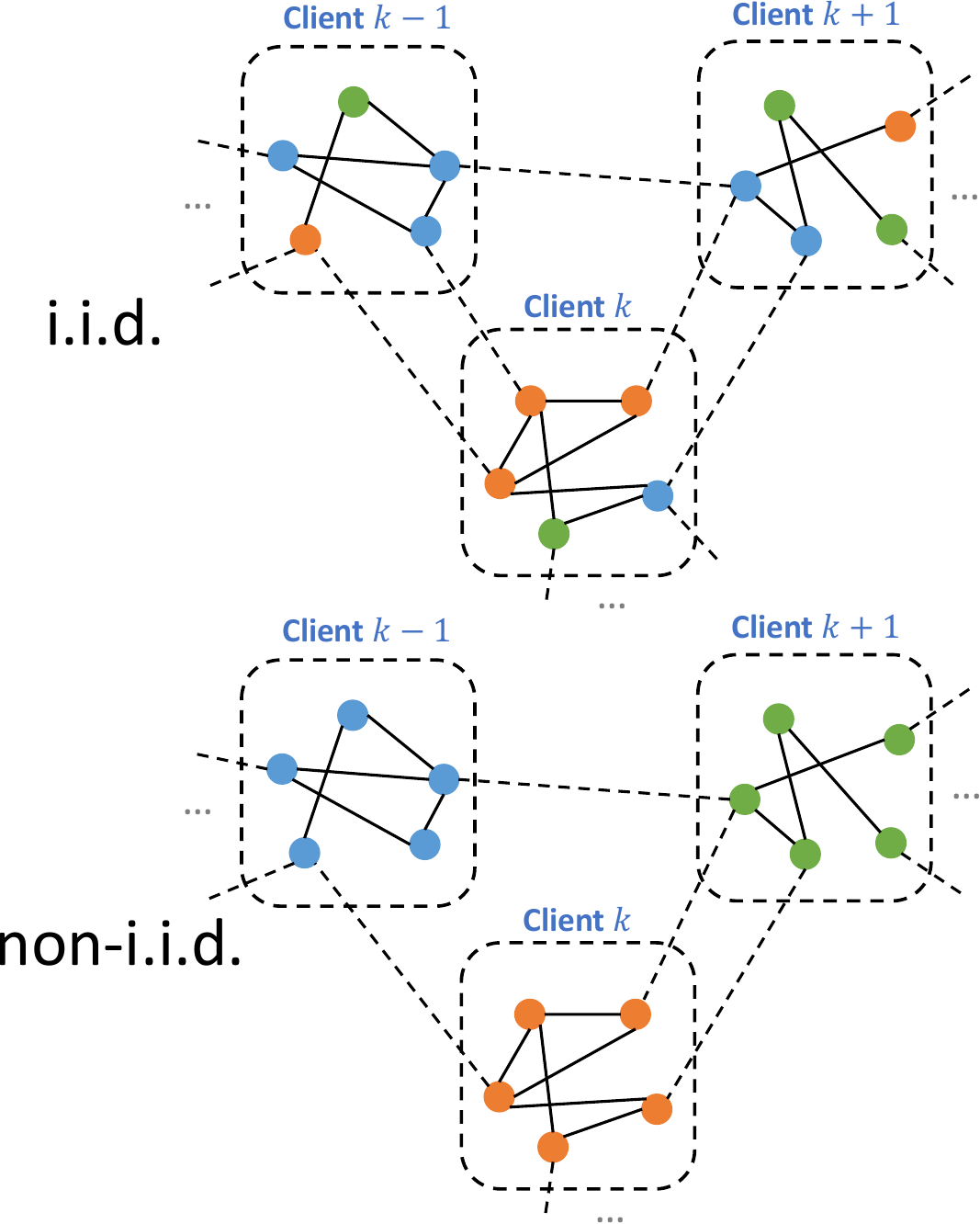}
    \caption{Federated GCN training schematic for node classification, with colors indicating the known labels of some nodes. Nodes in a graph (shown as circles) are distributed across clients, and dashed lines show cross-client edges between nodes at different clients. Arrows in the left figure indicate that each client can exchange updates with a central server during the training process to predict the unknown labels of the grey nodes in each client. At right, we show a graph with i.i.d (top) and non-i.i.d. (bottom) data distribution across clients, which affects the number of cross-client edges.}
    \label{fig:problem}
    %\vspace{-0.5cm}
\end{figure}

Graph convolutional networks (GCNs) have been widely used for applications ranging from fake news detection in social networks to anomaly detection in sensor networks~\citep{benamira2019semi,zhang2020semi}. This data, however, can be too large to be trained on a single server, e.g., records of billions of users' website visits. Strict data protection regulations such as the General Data Protection Regulation (GDPR) in Europe and Payment Aggregators and Payment Gateways (PAPG) in India also require that private data only be stored in local clients. %: for example, users in a social network may not want to reveal the websites they have visited. 
In non-graph settings, federated learning has recently shown promise for %preserving user privacy while training accurate models
training models on data that is kept at multiple clients~\citep{zhao2018federated,yang2021achieving}. Some papers have proposed federated training of GCNs~\citep{he2021fedgraphnn,zhang2021subgraph}. Typically, these consider a framework in which each client has access to a subset of a large graph, and clients iteratively compute local updates to a semi-supervised model on their local subgraphs, which are occasionally aggregated at a central server. Figure~\ref{fig:problem}(left) illustrates the federated node classification task of predicting unknown labels of local nodes in each client.

The main challenge of applying federated learning to GCN training tasks involving a single large graph is that \emph{cross-client edges exist among clients}. In Figure~\ref{fig:problem}, for example, we see that some edges will connect nodes in different clients. We refer to these as ``cross-client edges''. 

Such cross-client edges typically are stored in both clients. Intuitively, this is due to the fact that edges are generated when nodes at clients interact with each other. Thus, the interaction record, though not personal node characteristics, is then naturally stored at both nodes, i.e., in both clients. For example, a graph may represent buying behaviors (edges) that exist between users (nodes) in two countries (clients). Users in one country want to buy items in another country. The records of these transactions between users in different countries (i.e., the cross-client edges) are then stored in both clients. Due to the General Data Protection Regulation, however, sensitive user information (node features including name, zip code, gender, birthday, credit card number, email address, etc.) cannot be stored in another country. Yet these cross-client edges cannot be ignored: including cross-country transactions (cross-client edges) is key for training models that detect international money laundering and fraud. Another example is online social applications like Facebook and LinkedIn. Users in different countries can build connections with each other (e.g., a person in the United States becoming Facebook friends with a person in China). The users in both the U.S. and China would then have a record of this friendship link, while \emph{the personal user information cannot be shared across countries}.

However, GCNs require information about a node's neighbors to be aggregated in order to construct an embedding of each node that is used to accomplish tasks such as node classification and link prediction. Ignoring the information from neighbors located at another client, as in prior federated graph training algorithms~\citep{wang2020graphfl, he2021spreadgnn}, may then result in less accurate models due to loss of information from nodes at other clients. %Sending the features of neighboring nodes to other clients, however, { can introduce significant communication overhead and reveal private node information to other clients}~\citep{wan2022bns}.
%risks revealing private information to them.

Prior works on federated or distributed graph training reduce cross-client information loss by communicating information about nodes' neighbors at other clients in each training round~\citep{scardapane2020distributed, wan2022bns, zhang2021subgraph}, which can introduce significant communication overhead and reveal private node information to other clients. We instead realize that \emph{the information needed to train a GCN only needs to be communicated once}, before training. 
This insight allows us to further \emph{alleviate the privacy challenges of communicating node information between clients}~\citep{zhang2021subgraph}. 
%Moreover, each node at a given client only needs to know the \emph{accumulated} information about that node's neighbors at each other client, allowing us to communicate only the accumulated information. In practice, each client may contain several node neighbors, e.g., clients might represent a company's supply chain or social network data in different countries, which cannot leave the country due to privacy regulations. Each client would then receive aggregated information about all of a node's neighbors in a different country. If such aggregation is insufficient to preserve node privacy, 
Specifically, we leverage Homomorphic Encryption (HE), which can preserve client privacy in federated learning but introduces significant overhead for each communication round; with only one communication round, this overhead is greatly reduced. Further, in practice each client may contain several node neighbors, e.g., clients might represent social network data in different countries, which cannot leave the country due to privacy regulations. Each client would then receive aggregated feature information about all of a node's neighbors in a different country, which itself can help preserve privacy through accumulation across multiple nodes. In the extreme case when nodes only have one cross-client neighbor, we can further integrate differential privacy techniques~\citep{wei2020federated}. We propose the \textbf{FedGCN algorithm} for distributed GCN training based on these insights. \emph{FedGCN greatly reduces communication costs and speeds up convergence without information loss, compared with existing distributed settings}~\citep{scardapane2020distributed, wan2022bns, zhang2021subgraph}

%In this work, we observe that training a GCN \emph{does not require complete information about other nodes' features}. Instead, each node at a given client only needs to know the \emph{accumulated} information about that node's neighbors at each other client. In practice, there may be many more nodes than clients, so each client would receive information accumulated over multiple nodes. For example, clients might represent a company's supply chain or social network data in different countries, which cannot leave the country due to privacy regulations.
%universities and nodes students in a social network that crosses institutions. 
%Moreover, since this accumulated information does not depend on the GCN model, \emph{we need only communicate it once}, before training the GCN. We propose the \textbf{FedGCN algorithm} for distributed GCN training based on this insight. FedGCN greatly reduces communication costs and speeds up convergence without information loss, compared with existing distributed settings that reduce cross-client information loss with communication in each training round~\citep{scardapane2020distributed, wan2022bns, zhang2021subgraph}

% \begin{figure}[ht]
%     \centering
%     \includegraphics[width = 0.47\textwidth]{iid_non_iid_example.pdf}
%     \caption{Cross-client graph with i.i.d (left) and non-i.i.d. (right) data distribution. Node color represents the class of the node. Data distribution affects the number of cross-client edges.}
%     \label{fig:iid_non_iid_example}
% \end{figure}

In some settings, we can further reduce FedGCN's required communication without compromising the trained model's accuracy.
First, GCN models for node classification rely on the fact that nodes of the same class will have more edges connecting them, as shown in Figure~\ref{fig:problem}(right). If nodes with each class tend to be concentrated at a single client, a version of the non-i.i.d. (non-independent and identically distributed) data often considered in federated learning, then ignoring cross-client edges discards little information, and FedGCN's communication round may be unnecessary. The model, however, may not converge, as federated learning may converge poorly when client data is non-i.i.d.~\citep{zhao2018federated}. Second, GCN models with multiple layers require accumulated information from nodes that are multiple hops away from each other, introducing greater communication overhead. However, such multi-hop information may not be needed in practice. %It is thus unclear under which circumstances more communication is worth the convergence improvement.

We analytically quantify the convergence rate of FedGCN with various degrees of communication, under both i.i.d. and non-i.i.d. client data. To the best of our knowledge, we are the first to \emph{analytically illustrate the resulting tradeoff between a fast convergence rate (which intuitively requires more information from cross-client edges) and low communication cost}, which we do by considering a stochastic block model~\citep{lei2015consistency,keriven2020convergence} of the graph topology. We thus quantify when FedGCN's communication significantly accelerates the GCN's convergence.
%analysis the performance drop for block matrix setting on IID and Non-IID case, then propose Federated Graph Convolutional Network, an efficient framework for federated training which only needs communication of features at the initial step. Compared with distributed setting~\cite{scardapane2020distributed} which requires communication at each round, our method greatly reduces communication cost and speed up the convergence rate without information loss. We then analysis the convergence rate and communication cost. 
%\subsection{Our Contribution}

%Communicating data between clients additionally runs the risk of compromising privacy. FedGCN's accumulation of node data may itself preserve privacy if multiple nodes' information is aggregated. Unlike prior works that also face privacy challenges due to communicating node information between clients~\citep{zhang2021subgraph}, we further guarantee privacy by leveraging Homomorphic Encryption (HE)~\citep{ni2021vertical} to secure the neighbor feature aggregation and model gradient aggregation. In the extreme case when nodes only have one cross-client neighbor, resulting in limited privacy from accumulating node features, FedGCN can further integrate optional differential privacy techniques~\citep{wei2020federated}.

In summary, our work has the following \textbf{contributions:}
\begin{itemize}
    \item We introduce FedGCN, an \emph{efficient framework} for federated training of GCNs to solve node-level prediction tasks where distributed training incurs high communication cost and privacy leakage, which also leverages Fully Homomorphic Encryption for enhanced privacy guarantees.
    \item We \emph{theoretically analyze} the convergence rate and communication cost of FedGCN compared to prior methods, as well as its dependence on the data distribution. We can thus quantify the usefulness of communicating different amounts of cross-client information. 
    \item Our \emph{experiments} on both synthetic and real-world datasets demonstrate that FedGCN outperforms existing distributed GCN training methods in most cases with exact model computation, higher accuracy, and orders-of-magnitude (e.g. $100\times$) less communication cost.
    %Our framework can be easily extended to analyze other federated graph training methods. % Experiments validate our theoretical analysis.
%    \item Counter intuitive idea: Non-IID improves the performance, low communication loss
%    \item Trade off among 2-order, 1-order, 0-order methods.
    
\end{itemize}
We outline related works in Section~\ref{sec:related} before introducing the problem of node classification in graphs in Section~\ref{sec:problem}. We then introduce FedGCN in Section~\ref{sec:fedgcn} and analyze its performance theoretically (Section~\ref{sec:theory}) and experimentally (Section~\ref{sec:experiment}) before concluding in Section~\ref{sec:conclusion}.

%% file: 2-related_work.tex
\section{Related Work}\label{sec:related}
\textbf{Graph neural networks}
%Different from euclidean data, graph data has a geometric structure that needs to learn a high dimensional representation of nodes in the graph. I would rephrase this: 
aim to learn representations of graph-structured data~\citep{bronstein2017geometric}. GCNs~\citep{kipf2016semi}, GraphSage~\citep{hamilton2017inductive}, and GAT~\citep{velivckovic2017graph} perform well on node classification and link prediction.
Several works provide a theoretical analysis of GNNs based on the Stochastic Block Model~\citep{zhou2019analysis, lei2015consistency,keriven2020convergence}. We similarly adopt the SBM to quantify FedGCN's performance.

\textbf{Federated learning}
was first proposed in~\citet{mcmahan2017communication}'s widely adopted FedAvg algorithm, which allows clients to train a model via coordination with a central server while keeping training data at local clients. 
%Instead of uploading data to the server for centralized training, clients iteratively process their local data and share model updates with the server. Weights from a large population of clients are periodically aggregated by the server and combined to create an improved global model. 
%FedAvg and most of its variants use a central server to periodically combine local client updates into a new global model. 
However, FedAvg may not converge if data from different clients is non-i.i.d.~\citep{zhao2018federated, li2019convergence, yang2021achieving}. We show similar results for federated graph training.

\textbf{Federated learning on graph neural networks}
is a topic of recent interest~\citep{he2021fedgraphnn}. Unlike learning tasks in which multiple graphs each constitute a separate data sample and are distributed across clients (e.g., graph classification~\citep{zhang2018end}, image classification~\citep{li2019deepgcns}, and link prediction~\citep{yao2023fedrule}, 
%GraphFL~\citep{wang2020graphfl} is a model-agnostic meta learning approach, while ASFGNN~\citep{zheng2021asfgnn} is a Bayesian optimization technique to automatically tune the hyper-parameters of all clients for separated graphs. 
%
FedGCN instead considers semi-supervised tasks on a \emph{single large graph} (e.g., for node classification). Existing methods for such tasks generally ignore the resulting cross-client edges~\citep{he2021fedgraphnn}. 
%CNFGNN~\citep{meng2021cross} uses the central server to deal with spatial dependencies between nodes, but this method affects the privacy of users by revealing information about their data to the server. 
\citet{scardapane2020distributed}'s distributed GNN proposes a training algorithm communicating the neighbor features and intermediate outputs of GNN layers among clients with expensive communication costs. BDS-GCN~\citep{wan2022bns} proposes to sample cross-client neighbors. These methods may violate client privacy by revealing per-node information to other clients. FedSage+~\citep{zhang2021subgraph} recovers missing neighbors for the input graph based on the node embedding, which requires fine-tuning a linear model of neighbor generation and may not fully recover the cross-client information. It is further vulnerable to the data reconstruction attack, compromising privacy.

%
%state it is a concurrent work

All of the above works further require communication at every training round, while FedGCN enables the private recovery of cross-client neighbor information with a \emph{single}, pre-training communication round that utilizes HE. We also provide theoretical bounds on FedGCN's convergence.

%a theoretical analysis on the convergence of FedGCN, which may also be used to analyze the convergence of these previously proposed algorithms and can thus help to compare their performance.
%
%
%BDS-GCN: Boundary Sampling
%
%Theoretical analysis on convergence rate is also needed to better evaluate the algorithms. Our work is to fill the gap.

%% file: 3-problem_statement.tex
\section{Federated Semi-Supervised Node Classification}\label{sec:problem}
In this section, we formalize the problem of node classification on a single graph and introduce the federated setting in which we aim to solve this problem. %In Section~\ref{sec:gcn}, we describe the GCN model used to solve this problem.

%\subsection{Node Classification in Centralized Learning}\label{subsec:central}
We consider a graph $\mathcal{G} = (\mathcal{V}, \mathcal{E})$, where $\mathcal{V} = [N]$ is the set of $N$ nodes and $\mathcal{E}$ is the set of edges. The graph is equivalent to a weighted adjacency matrix $\bm{A}\in \mathbb{R}^{N\times N}$, where $\bm{A}_{ij}$ indicates the weight of an edge from node $i$ to node $j$ (if the edge does not exist, the weight is zero). Every node $i\in \mathcal{V}$ has a feature vector $\bm{x}_i \in \mathbb{R}^d$, where $d$ represents the number of input features. Each node $i$ in a subset $\mathcal{V}^{train}\subset \mathcal{V}$ has a corresponding label $y_i$ used during training. Semi-supervised node classification aims to assign labels to nodes in the remaining set $\mathcal{V} \backslash \mathcal{V}^{train}$, based on their feature vectors and edges to other nodes.  We train a GCN model to do so.

GCNs~\citep{kipf2016semi} consist of multiple convolutional layers, each of which constructs a node embedding by aggregating the features of its neighboring nodes. Typically, the node embedding matrix $\bm{H}^{(l)}$ for each layer $l = 1,2,\ldots,L$ is initialized to $\bm{H}^{(0)}=\bm{X}$, the matrix of features for each node (i.e., each row of $\bm{X}$ corresponds to the features for one node), and follows the propagation rule
%A multi-layer Centralized Graph Convolutional Network (GCN)~\cite{kipf2016semi} has the layer-wise propagation rule:
%\begin{equation}
   $\bm{H}^{(l+1)}=\phi(\bm{A} \bm{H}^{(l)}\bm{W}^{(l)})$.
%\end{equation}
Here $\bm{W}^{(l)}$ are parameters to be learned, $\bm{A}$ is the weighted adjacency matrix, and $\phi$ is an activation function. Typically, $\phi$ is chosen as the softmax function in the last layer, so that the output can be interpreted as the probabilities of a node lying in each class, with ReLU activations in the preceding layers.
%The adjacency matrix $\bm{A}$ typically requires adding self loop and normalization to have better performance. The activation function is $\phi$. The node embedding matrix in the $l$-th layer is $\bm{H}^{(l)}\in \mathbb{R}^{N\times D}$, which contains high-level representations of the graph nodes transformed from the initial features; $H^{(0)}=\bm{X}$. 
%
The embedding of each node $i\in \mathcal{V}$ at layer $l+1$ is then
\begin{equation}\label{equ:gcn_node_central}
    \bm{h}^{(l+1)}_i=\phi\left(\sum_{j\in\mathcal{N}_{i}} \bm{A}_{ij}\bm{h}_j^{(l)}\bm{W}^{(l)}\right),
\end{equation}
which can be computed from the previous layer's embedding $\bm{h}_j^{(l)}$ for each neighbor $j$ and the weight $\bm{A}_{ij}$ on edges from node $i$ to node $j$.
For a GCN with $L$ layers in this form, the output for node $i$ will depend on neighbors up to $L$ steps away (i.e., there exists a path of no more than $L$ edges to node $i$). We denote this set by $\mathcal{N}_i^L$ (note that $i\in\mathcal{N}_i^L$) and refer to these nodes as $L$-hop neighbors of $i$.

%\subsection{Node Classification in Federated Learning}\label{subsec:federated}
%Mobile devices

To solve the node classification problem in \textbf{federated settings} (Figure \ref{fig:problem}), we consider, as usual in federated learning, a central server with $K$ clients. The graph $\mathcal{G} = (\mathcal{V}, \mathcal{E})$ is separated across the $K$ clients, each of which has a sub-graph $\mathcal{G}_k = (\mathcal{V}_k, \mathcal{E}_k)$. Here $\bigcup_{k=1}^K \mathcal{V}_k = \mathcal{V} $ and $ \mathcal{V}_i \bigcap \mathcal{V}_j = \varnothing$ for $\forall i\neq j \in [K] $, i.e., the nodes are disjointly partitioned across clients. The features of nodes in the set $\mathcal{V}_k$ can then be represented as the matrix $\bm{X}_k$. The cross-client edges of client $k$, $\mathcal{E}_k^c$, for which the nodes connected by the edge are at different clients, are known to the client $k$.
%The set of cross-client edges, for which the nodes connected by the edge are at different clients, is denoted by $\mathcal{E}_C = \mathcal{E} \backslash \bigcup_{k=1}^K \mathcal{E}_k$. 
We use $\mathcal{V}_k^{train} \subset \mathcal{V}_k$ to denote the set of training nodes with associated labels $y_i$. The task of federated semi-supervised node classification is then to assign labels to nodes in the remaining set $\mathcal{V}_k \backslash \mathcal{V}_{k}^{train}$ for each client $k$.

%Every node is associated with a node feature vector $\bm{x_i} \in \mathbb{R}^d$. 
%The edges between nodes in client $i$ and client $j$ is denoted by $\mathcal{E}_{i,j}$, where  $ \bigcup_{i,j \in [K]} \mathcal{E}_{i,j}= \mathcal{E} \backslash \bigcup_{k=1}^K \mathcal{E}_k$.
%\mathcal{E}_{i,j} is known for clients i and j
%In each client $k$, every node $i$ in a subset $\mathcal{V}_{k}^{train}\subset \mathcal{V}_k$ of nodes is associated with a label $y_i$, which is used in the training. The task of semi-supervised node classification is to assign labels to nodes in the remaining set $\mathcal{V}_k \backslash \mathcal{V}_{k}^{train}$. 

%Applying GCNs in the federated setting immediately raises a challenge. 
As seen from (\ref{equ:gcn_node_central}), in order to find the embedding of the $i$-th node in the $l$-th layer, we need the previous layer's embedding $\bm{h}_j^{(l)}$ for all neighbors of node $i$. In the federated setting, however, some of these neighbors may be located at other clients, and thus their embeddings must be iteratively sent to the client that contains node $i$ for each layer at every training round. \citet{he2021fedgraphnn} ignore these neighbors, considering only $\mathcal{G}_k$ and $\mathcal{E}_k$ in training the model, while~\citet{scardapane2020distributed,wan2022bns,zhang2021subgraph} require such communication, which may lead to high overhead and privacy costs. FedGCN provides a communication-efficient method to account for these neighbors.
%The main challenge is that: For each client $k$, the node feature $\bm{X}_k$ and edges $\mathcal{E}_k$ are stored locally. We only assume nodes in each client knows their neighbor nodes stored in other clients, which means $\mathcal{E}_D$ is partially known to each client. The node features in other clients are unknown to the client.

%Simply only consider $\mathcal{E}_k$ and $\bm{X}_k$ like in former works \cite{wang2020graphfl,zheng2021asfgnn} without $\mathcal{E}_D$ and node features in other clients causes great information loss, which makes model hard to converge with insufficient classification performance.
%need a figure to show the information loss

%In order to overcome information loss, former work~\cite{scardapane2020distributed} requires communication at each iteration, which causes great communication cost.

%An optimized convergence and communication tradeoff is needed.

%% file: 4-fedgcn.tex
\section{Federated Graph Convolutional Network}\label{sec:fedgcn}

In order to overcome the challenges outlined in Section~\ref{sec:problem}, we propose our Federated Graph Convolutional Network (FedGCN) algorithm. In this section, we first introduce our federated training method with communication at the initial step and then outline the corresponding training algorithm.

\textbf{Federating Graph Convolutional Networks.}\label{subsec:federated-gcn} In the federated learning setting, let $c(i)$ denote the index of the client that contains node $i$ and $\bm{W}^{(l)}_{c(i)}$ denote the weight matrix of the $l$-th GCN layer of client $c(i)$. The embedding of node $i$ at layer $l+1$ is then $\bm{h}^{(l+1)}_i=\phi\left(\sum_{j\in\mathcal{N}_{i}} \bm{A}_{ij}\bm{h}^{(l)}_j \bm{W}^{(l)}_{c(i)}\right)$.

% \begin{equation}\label{equ:gcn_node_i}
% %    h^{(l+1)}_i=\phi(\sum_{j\in\mathcal{N}_{i}} \bm{A}_{ij}h^{(l)}_j\bm{W}^{(l)}_{c(i)}),
% \bm{h}^{(l+1)}_i=\phi\left(\sum_{j\in\mathcal{N}_{i}} \bm{A}_{ij}\bm{h}^{(l)}_j \bm{W}^{(l)}_{c(i)}\right).
% \end{equation}
Note that the weights $\bm{W}^{(l)}_{c(i)}$ may differ from client to client, due to the independent local training in federated learning.
%
%In practice, GCNs often require only two or three layers for node-level prediction tasks (node classification and link prediction)~\citep{kipf2016semi} to have sufficient performance. We experimentally validate the need for a limited number of layers in the paper's appendix. 
For example, we can then write the computation of a 2-layer federated GCN as
%\begin{equation}
    $\bm{\hat{y}}_i=\phi\left(\sum_{j\in\mathcal{N}_i}\bm{A}_{ij}\phi\left(\sum_{m\in\mathcal{N}_j}\bm{A}_{jm}\bm{x}_m^T \bm{W}^{(1)}_{c(i)}\right)\bm{W}^{(2)}_{c(i)}\right)$.
%\end{equation} 
To evaluate this 2-layer model, it then suffices for the client $k=c(i)$ to
%Based on this idea, the clients can first communicate the information of nodes. After communication of information, we can then train the model. We then introduce a way of communication without information loss but keep the privacy. 
%
%
%To evaluate this at node $i$, the client $k=c(i)$ must 
receive the message $\sum_{m\in\mathcal{N}_j}\bm{A}_{jm}\bm{x}_m^T$. We can write these messages as
\begin{equation}
    \sum_{j\in \mathcal{N}_i} \bm{A}_{ij}  \bm{x}_j,\,\text{and}\,
%    \label{equ:message1} \\
%\end{equation}
%
%\begin{equation}
    \left\{\sum_{m\in \mathcal{N}_j} \bm{A}_{jm}  \bm{x}_m\right\}_{j\in \mathcal{N}_{i} / i},
    \label{equ:message}
\end{equation}
which are the feature aggregations of 1- and 2-hop neighbors of node $i$ respectively. \emph{This information does not change with the model training}, as it simply depends on the (fixed) adjacency matrix $\bm{A}$ and node features $\bm{x}$. The client also naturally knows $\{\bm{A}_{ij}\}_{\forall j\in \mathcal{N}_i}$, which is included in $\mathcal{E}_k \bigcup \mathcal{E}_k^c$. 

One way to obtain the above information is to receive the following message from clients $z$ that contain at least one two-hop neighbor of $k$:
%$z\in \mathcal{C}_i^1$:
\begin{equation}
    \sum_{j\in \mathcal{N}_i} \mathbb{I}_z (c(j)) \bm{A}_{ij}  \bm{x}_j,\,\text{and}\,
%    \label{equ:message1} \\
%\end{equation}
%
%\begin{equation}
    \forall j\in \mathcal{N}_i, \sum_{m\in \mathcal{N}_j} \mathbb{I}_z (c(m))\cdot \bm{A}_{jm}  \bm{x}_m.
    \label{equ:p2pmessage}
\end{equation}

%$ \{\sum_{j\in \mathcal{N}_i} \mathbb{I}_z (c(j)) \bm{A}_{ij}  \bm{x}_j\}_{z\in[K]}$ \carlee{what is this stray equation?}

Here the indicator $\mathbb{I}_z (c(m))$ is 1 if $z = c(m)$ and zero otherwise. More generally, for a $L$-layer GCN, each layer requires $\forall j\in \mathcal{N}_i^L / \mathcal{N}_i^{L-1}, \sum_{m\in \mathcal{N}_j} \mathbb{I}_z (c(m))\cdot \bm{A}_{jm}  \bm{x}_m$.
%
% \begin{equation}
%     \forall j\in \mathcal{N}_i^L / \mathcal{N}_i^{L-1}, \sum_{m\in \mathcal{N}_j} \mathbb{I}_z (c(m))\cdot \bm{A}_{jm}  \bm{x}_m.
%     \label{equ:message3}
% \end{equation}
%
Further, $\mathcal{E}_i^{L-1}$, i.e., the set of edges up to $L-1$ hops away from node $i$, is needed for normalization of $A$. %However, this method requires communication among multiple pairs of clients and suffers privacy leakage in a case when there is only one neighbor node in the client. % \yuhang{Here need a figure for showing server aggregation}

To avoid the overhead of communicating between multiple pairs of clients, which can also induce privacy leakage when there is only one neighbor node in the client, we can instead send the aggregation of each client to the central server. In the 2-layer GCN example, the server then calculates the sum of neighbor features of node $i$ as
%\begin{equation}
    $\sum_{j\in \mathcal{N}_i} \bm{A}_{ij}  \bm{x}_j  = \sum_{k=1}^K \sum_{j\in \mathcal{N}_i} \mathbb{I}_k (c(j))\cdot \bm{A}_{ij}  \bm{x}_j$.
%\end{equation}
%To normalize $A$, a popular technique for training GCNs, the client must also know $A_{im}, \forall i\in \mathcal{N}_m$, which is included in $E_k$. 
%
The server can then send the required feature aggregation in (\ref{equ:message}) back to each client $k$. 
Thus, we only need to send the \emph{accumulated} features of each node's (possibly multi-hop) neighbors, in order to evaluate the GCN model. If there are multiple neighbors stored in other clients, this accumulation serves to protect their individual privacy\footnote{In the extreme case when the node only has one neighbor stored in other clients, we can drop the neighbor, which likely has minimal effect on model performance, or add differential privacy to the communicated data.}. For the computation of all nodes $\mathcal{V}_k$ stored in client $k$ with an $L$-layer GCN, the client needs to receive 
$
    \{\sum_{j\in \mathcal{N}_i} \bm{A}_{ij}  \bm{x}_j\}_{i\in \mathcal{N}_{\mathcal{V}_k}^L},
$
where $\mathcal{N}_{\mathcal{V}_k}^L$ is the set of $L$-hop neighbors of nodes $\mathcal{V}_k$.

%Based on this trick, we only need to communicate the accumulated node features with privacy preservation. 
FedGCN is based on the insight that GCNs require only the accumulated information of the $L$-hop neighbors of each node, which may be communicated in advance of the training. In practice, however, even this communication may be infeasible. %For example, in the extreme case of each client consisting of a single node (such as a user in a social network), accumulating features at different clients does not offer any privacy benefits. 
If $L$ is too large, $L$-hop neighbors may actually consist of the entire graph (social network graphs have diameters $< 10$), which might introduce prohibitive storage and communication requirements. Thus, we design FedGCN to accommodate three types of communication approximations, according to the most appropriate choice for a given application:
%Given the privacy concern and communication cost given the $\mathcal{E}_i^{L-1}$, we then introduce 3-types of approximate algorithm without knowing $\mathcal{E}_i^{L-1}$.
%\subsubsection{0-order Approximation}\label{subsubsec:0-order}

\begin{itemize}

\item \textbf{No communication (0-hop):} If any communication is unacceptable, e.g., due to overhead, each client simply trains on $\mathcal{G}_k$ and ignores cross-client edges, as in prior work.
%Block matrix. Without communication.

%\subsubsection{1-order Approximation}\label{subsubsec:1-order}
\item \textbf{One-hop communication:} If some communication is permissible, we may use the accumulation of feature information from nodes' 1-hop neighbors, $\{\sum_{j\in \mathcal{N}_i} \bm{A}_{ij}  \bm{x}_j\}_{i\in {\mathcal{V}_k}}$, to approximate the GCN computation. 1-hop neighbors are unlikely to introduce significant memory or communication overhead as long as the graph is sparse, e.g. social networks.
%Compared to including all $L$-hop neighbors, this method reduces the number of neighbors that need to send information by only requiring the communication of (\ref{equ:message1}). The features corresponding to each node's one-hop neighbors are then included in the GCN.
%\tdCarlee{this should be explained better}
% \begin{equation}
%     \sum_{l\in \mathcal{N}_i} \mathbb{I}_z (c(l))\cdot A_{il}  x_l.
%     \label{equ:message1}
% \end{equation}
%\subsubsection{2-order Approximation}\label{subsubsec:2-order}
\item \textbf{Two-hop communication:} To further improve model performance,  we can communicate the information from 2-hop neighbors, $\{\sum_{j\in \mathcal{N}_i} \bm{A}_{ij}  \bm{x}_j\}_{i\in \mathcal{N}_{\mathcal{V}_k}}$ and perform the aggregation of $L$-layer GCNs. 
%For a 2-layer GCN, communicating the information can perfectly recover all neighboring nodes' information (i.e., there is no information loss). 
As shown in Figure~\ref{fig:centralized_fedgcn_comparison}, the 2-hop approximation does not decrease model accuracy in practice compared to $L$-hop communication for $L$-hop GCNs, up to $L\leq 10$. 
  %For a 2-layer GCN, communicating the information can perfectly recover all neighboring nodes' information (i.e., there is no information loss). %This choice requires more communication than one-hop communication as there are more two-hop neighbors than one-hop neighbors. 
\begin{figure}[ht]
    \centering
    \includegraphics[width = 0.33\textwidth]{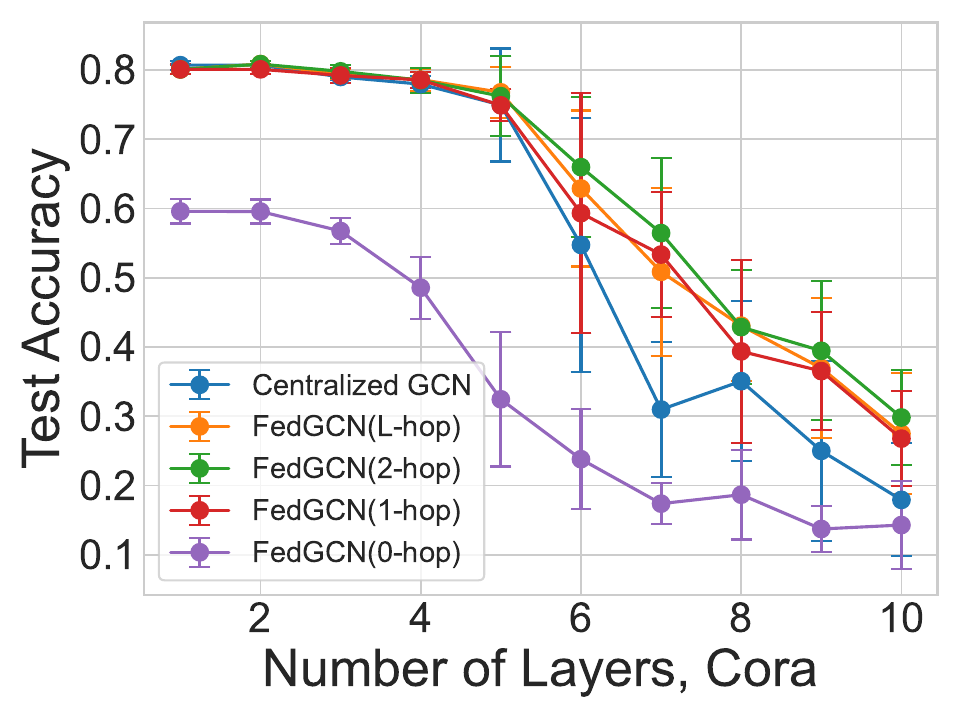}%0.38
    \hspace{0.03\textwidth}
    \includegraphics[width = 0.33\textwidth]{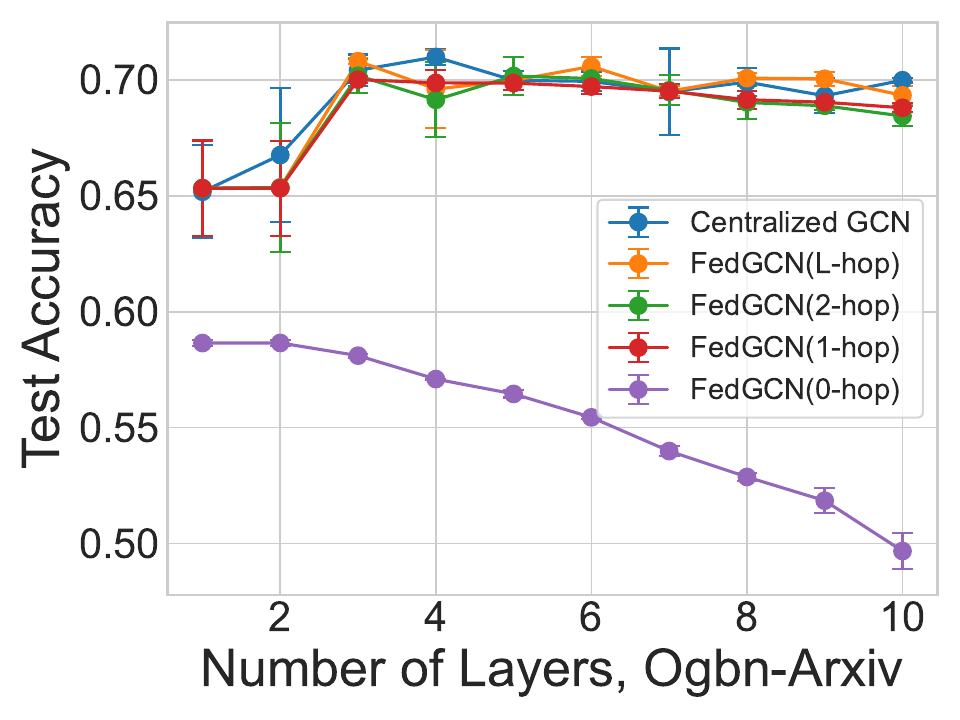}
    \caption{Test Accuracy of GCNs with different numbers of layers, using centralized and FedGCN training on Cora (left) and Ogbn-Arxiv (right) datasets with 10 clients and non-i.i.d partition. Communicating 2-hop information in FedGCN is sufficient for up to 10-layer GCNs. In Ogbn-Arxiv, BatchNorm1d is added between GCN layers to ensure consistent performance~\citep{hu2020open}.}
    \label{fig:centralized_fedgcn_comparison}
    \vspace{-0.3cm}
\end{figure}
\end{itemize}
\textbf{Why Not $L$-hop Communication?} Although FedGCN supports $L$-hop communication, communicating $L$-hop neighbors across clients requires knowledge of the $L - 1$ hop neighborhood graph structures, which may incur privacy leakage. If $L$ is large enough, $L$-hop neighbors may also cover the entire graph, incurring prohibitive communication and computation costs. Thus, in practice we restrict ourselves to 2-hop communication, which only requires knowledge of 1-hop neighbors. Indeed, Figure~\ref{fig:centralized_fedgcn_comparison} shows that even on the Ogbn-Arxiv dataset, which has more than one million edges, adding more layers and $k$-hop communication does not increase model accuracy for $L \geq 3$ and $k\geq 2$. Thus, it is reasonable to use 0, 1, or 2-hop communication with $L\leq 3$-layer GCNs in practice.
%will cover the whole global graph, which incurs prohibitive communication and computation cost. The improvement of the model performance is also minor. L-hop communication also requires known L-1 hop neighborhood graph structures, which may incur privacy leakage. How to utilize the L-1 hop neighborhood graph structures in a privacy-preserving way is still an open problem. Also, the GNN precision performance~\citep{kipf2016semi, hu2020open} drops when the number of layers increases.

\textbf{Secure Neighbor Feature Aggregation.} To guarantee privacy during the aggregation process of accumulated features, we leverage Homomorphic Encryption (HE) to construct a secure neighbor feature aggregation function. HE~\citep{brakerski2014leveled, cheon2017homomorphic} allows a computing party to perform computation over ciphertext without decrypting it, thus preserving the plaintext data. 
%The values of the neighbor features are generally integers whereas the model parameters are usually real numbers. Due to this property, we select the BGV variant and the CKKS variant of FHE respectively as the building component of our secure function.

%This also means aggregating $n$ neighbor feature arrays with an array size of $m$ requires a complexity of $(n-1)*m$.
The key steps of the process can be summarized as follows:
%(integrated with the main framework in Algorithm \ref{Alg:fedgcn}) 
(i) all clients agree on and initialize a HE keypair, (ii) each client encrypts the local neighbor feature array and sends it to the server, and (iii) upon receiving all encrypted neighbor feature arrays from clients, the server performs secure neighbor feature aggregation 
%\begin{equation}
    $\left[\!\left[\sum_{j\in \mathcal{N}_i} \bm{A}_{ij}  \bm{x}_j \right]\!\right] = \sum_{k=1}^K \left[\!\left[ \sum_{j\in \mathcal{N}_i} \mathbb{I}_k (c(j))\cdot \bm{A}_{ij}  \bm{x}_j\right]\!\right]$,
%\end{equation}
where $[\![\cdot]\!]$ represents the encryption function. The server then distributes the aggregated neighbor feature array to each client, and (iv) upon receiving the aggregated neighbor feature array, each client decrypts it and moves on to the model training phase.
%
%The aggregation server performs neighbor feature aggregation without having access (i.e., decrypting) to plaintext neighbor features from each client. Each client only receives the accumulation of neighbor features. Differential privacy methods can also be integrated in highly privacy-sensitive applications. %Additionally, with proper access control, clients will not have direct access to local neighbor features from other clients in the system.
We can also use HE for a secure model gradient aggregation function \emph{during the model's training rounds}, which provides extra privacy guarantees. %by restricting knowledge of local client models. 

Since model parameters are often floating point numbers and node features can be binary (e.g., one-hot indicators), a na\"ive HE scheme would use CKKS~\citep{cheon2017homomorphic} for parameters and integer schemes such as BGV~\citep{brakerski2014leveled} for features.
%, which requires the use of approximate HE scheme CKKS~\citep{cheon2017homomorphic}. While our neighbor feature aggregation is originally designed for integers with integer schemes such as BGV~\citep{brakerski2014leveled}, 
To avoid the resulting separate cryptographic setups, %(a key management procedure for both CKKS and BGV respectively), 
we adopt CKKS with a rounding procedure for integers and also propose an efficient HE file optimization, Boolean Packing, that packs arrays of boolean values into integers and optimizes the cryptographic communication overhead. The encrypted features then only require twice the communication cost of the raw data, compared to 20x overhead with general encryption. %(20$\times$ the communication cost of the plaintext data). Boolean Packing works by packing arrays of boolean values into integers.

\textbf{Training Algorithm.} Based on the insights in the previous section, we introduce the FedGCN training algorithm (details are provided in Appendix~\ref{appen:algorithm}):
%shown in Algorithm \ref{Alg:fedgcn}. 
\begin{itemize}
    \item \textbf{Pretraining Communication Round} The algorithm requires communication between clients and the central server at the initial communication round. 
    \begin{enumerate}
        \item Each client $k$ sends its encrypted accumulations of local node features,  $[\![\{\sum_{j\in \mathcal{N}_i} \mathbb{I}_k (c(j))\cdot \bm{A}_{ij}  \bm{x}_j\}_{i \in {\mathcal{V}_k}}]\!]$, to the server.
        \item The server then accumulates the neighbor features for each node $i$, $[\![\sum_{j\in \mathcal{N}_i} \bm{A}_{ij}  \bm{x}_j ]\!] = \sum_{k=1}^K [\![ \sum_{j\in \mathcal{N}_i} \mathbb{I}_k (c(j))\cdot \bm{A}_{ij}  \bm{x}_j]\!]$.
        \item Each client receives and decrypts the feature aggregation of its one-hop, $[\![\{\sum_{j\in \mathcal{N}_i} \bm{A}_{ij}  \bm{x}_j\}_{i\in {\mathcal{V}_k}}]\!]$, and if needed two-hop, neighbors $[\![\{\sum_{j\in \mathcal{N}_i} \bm{A}_{ij}  \bm{x}_j\}_{i\in \mathcal{N}_{\mathcal{V}_k}}]\!]$.
    \end{enumerate}

    \item \textbf{Federated Aggregation} After communication, FedGCN uses the standard FedAvg algorithm~\citet{mcmahan2017communication} to train the models. 
    %Specifically, each client computes $\tau$ gradient descent steps for local updates. Here we use $\bm{w}_k^{t,e}$ to denote the concatenation of the weights $W^{(l)}_k$ across the $L$ GCN layers, for client $k$ in global training round $t$ and local training step $e$, and $f_k$ to denote the local loss function, e.g., the cross entropy of the classification estimates. After $\tau$ local steps, the local model updates at clients are sent to the central server for the global model update, which is again secured through HE, and the new global model is pushed back to all clients to begin the next training round. The process repeats for $T$ global rounds until convergence. 
    Other federated learning methods, e.g., as proposed by ~\citet{reddi2020adaptive, fallah2020personalized}, can easily replace this procedure.%'s aggregation or local update procedures.
\end{itemize}
 
%the global model and it can be changed to other federated aggregation methods.

%% file: 5-theory.tex
\section{FedGCN Convergence and Communication Analysis}\label{sec:theory}
%Given our empirical findings on the convergence-communication tradeoff in Section~\ref{sec:experiment},
In this section, we theoretically analyze the convergence rate and communication cost of FedGCN for i.i.d. and non-i.i.d. data with 0-, 1-, and 2-hop communication. %We also empirically validate our analysis on the Stochastic Block Model~\citep{holland1983stochastic,abbe2017community} and real datasets. 

\begin{figure}[ht]
    \centering
    \includegraphics[trim={5mm 0 20mm 0},clip, width = 0.137\textwidth]{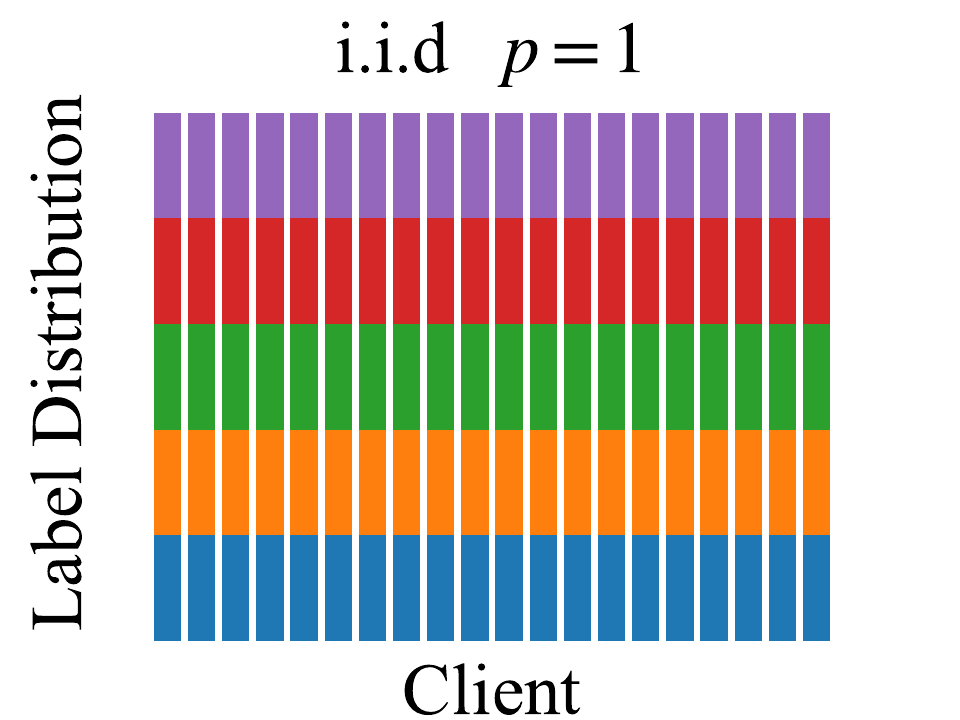}
    \includegraphics[trim={20mm 0 20mm 0},clip,width = 0.117\textwidth]{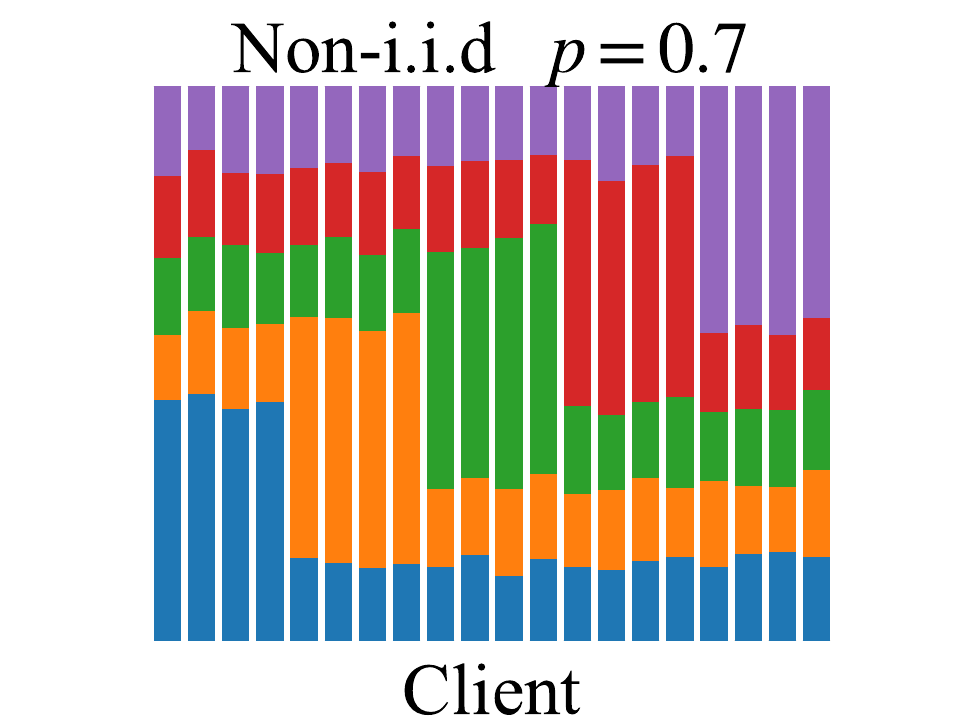}
    \includegraphics[trim={20mm 0 20mm 0},clip,width = 0.117\textwidth]{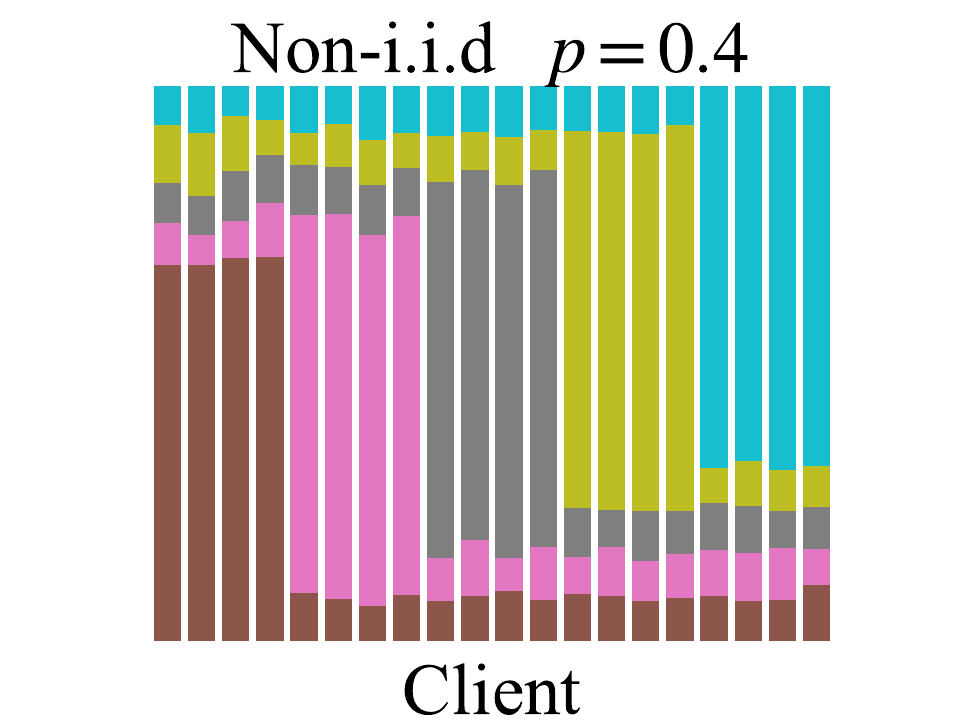}
    \includegraphics[trim={20mm 0 20mm 0},clip,width = 0.117\textwidth]{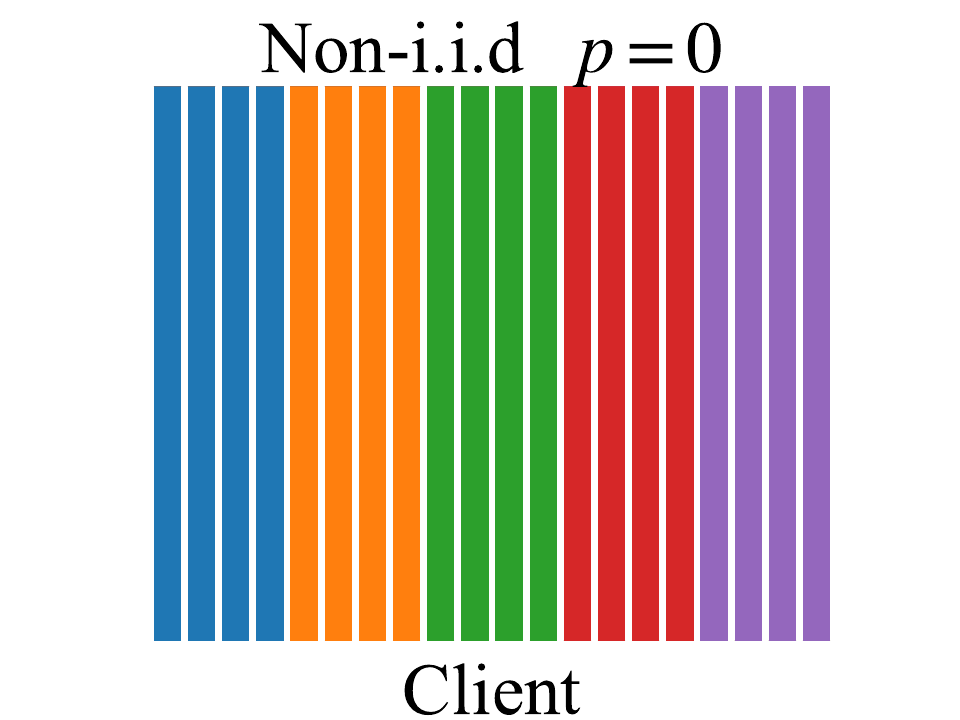}
    \includegraphics[trim={20mm 0 20mm 0},clip,width = 0.117\textwidth]{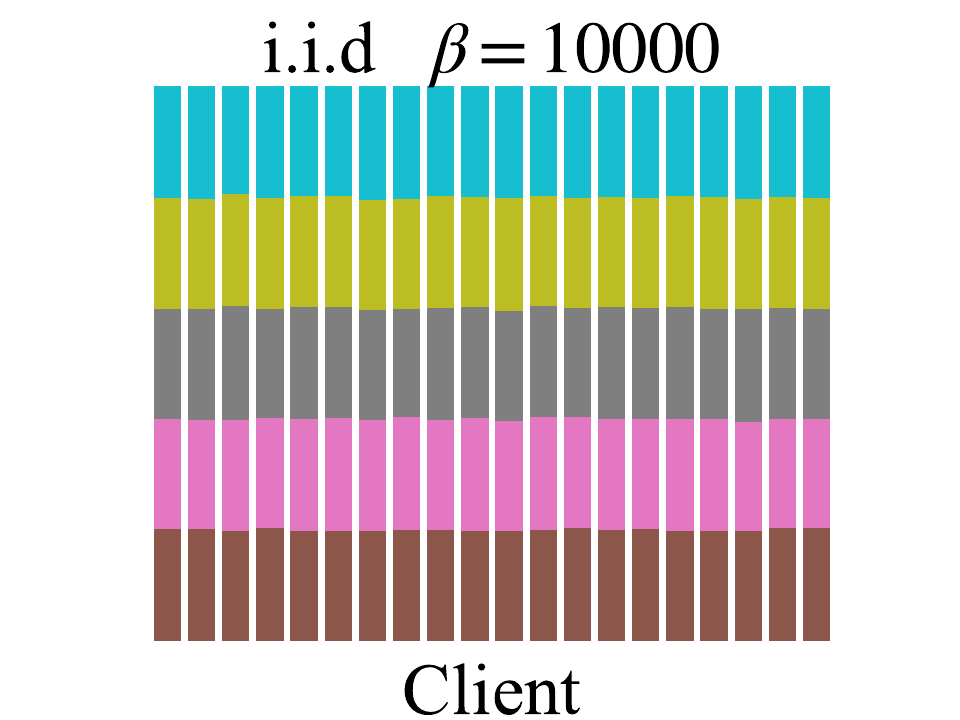}
    \includegraphics[trim={20mm 0 20mm 0},clip,width = 0.117\textwidth]{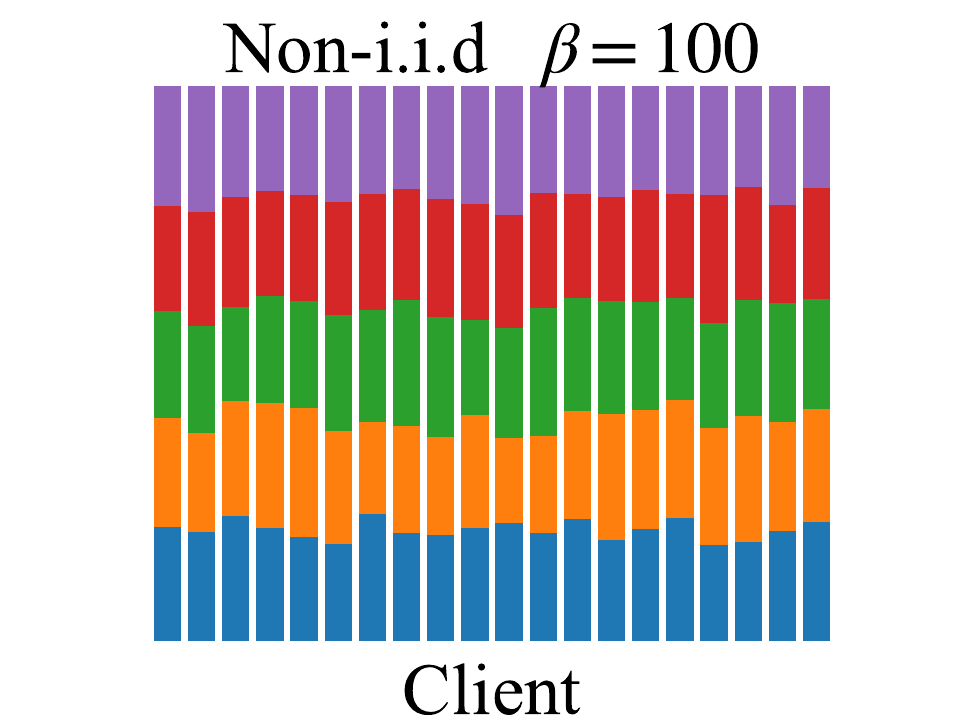}
    \includegraphics[trim={20mm 0 20mm 0},clip,width = 0.117\textwidth]{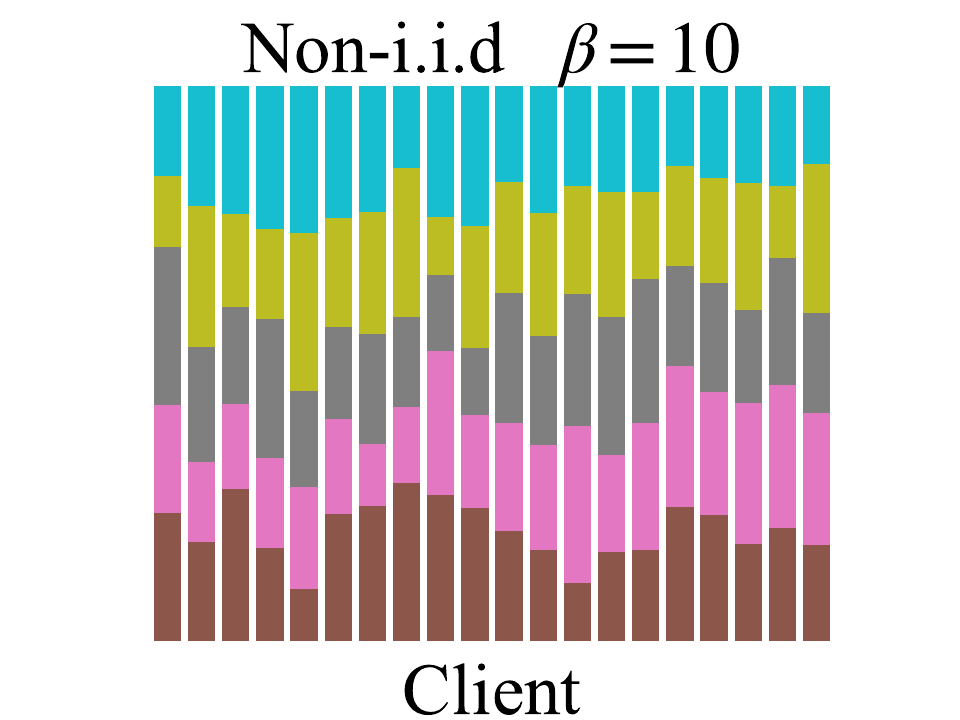}
    \includegraphics[trim={20mm 0 20mm 0},clip,width = 0.117\textwidth]{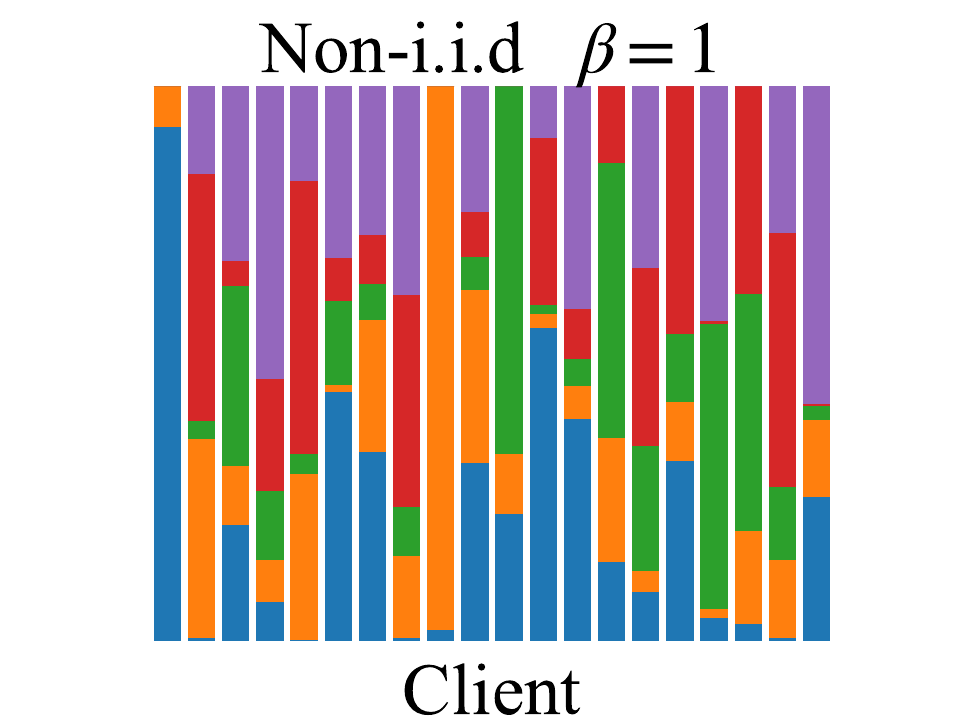}
    \vspace{-4mm}
    \caption{Example client data distributions. Different colors represent different node label classes. Client data is generated from the data distribution with different parameters $p$ and $\beta$ respectively.}
    \label{fig:clients_label_distribution}
    \vspace{-0.4cm}
\end{figure}

%\subsection{Definition of Data Distribution across Clients}
We first give a formal definition of the i.i.d. and non-i.i.d. data distributions, using distribution-based label imbalance~\citep{hsu2019measuring,li2022federated}. Figure~\ref{fig:clients_label_distribution} visualizes eight example data distributions across clients. For simplicity, we assume the number of clients $K$ exceeds the number of node label classes $C$, though Section~\ref{sec:experiment}'s experiments support any number of clients. We also assume that each class contains the same number of nodes and that each client has the same number of nodes.
\begin{definition}
Each client $k$'s \textit{label distribution} is defined as $\begin{bmatrix} p_1 & p_2 & \ldots & p_C\end{bmatrix}\in\mathbb{R}^C$, where $p_c$ denotes the fraction of nodes of class $c$ at client $k$ and $\sum_c p_c = 1$.
\end{definition}
\begin{definition}
Clients' data distributions are \textit{i.i.d.} when nodes are uniformly and randomly assigned to clients, i.e., each client's label distribution is $\begin{bmatrix}1/C & \ldots & 1/C\end{bmatrix}$. %The expected number of nodes for each label is then $[\frac{N}{KM}, ..., \frac{N}{K M}]^K$
% \end{definition}
%
% %\begin{definition}
%  %The expected number of nodes for each label is then $[\frac{N}{M}, 0, ..., 0]^K$
% %\end{definition}
%
% \begin{definition}
Otherwise, they are \textit{non i.i.d}.
%The expected number of nodes for each label is then $[(1-p + \frac{p}{K}) \frac{N}{M}, \frac{pN}{KM}, ..., \frac{pN}{KM}]^K$
\end{definition}

Non-i.i.d. distributions include that of~\cite{mcmahan2017communication} in which $p_i = 1 - p + \frac{p}{C}$ for some $i = 1,2,\ldots,C$ and $\frac{p}{C}$ otherwise, where $p\in [0,1]$ is a control parameter; or the Dirichlet distribution~\citep{hsu2019measuring} $Dir (\beta / C)$, where $\beta\geq 0$ is a control parameter. With these distributions, each client has one dominant class. If $p=0$ or $\beta \rightarrow \infty$, all nodes at a client have the same class.

%\subsection{Convergence Rate}

%\subsubsection{Preliminary}
%We next define some notation and assumptions. 
We use
$\|\bm{x}\|$ to denote the $\ell_2$ norm if $\bm{x}$ is a vector, and the Frobenius norm if $\bm{x}$ is a matrix. 
Given model parameters $\bm{w}$ and $K$ clients, we define the local loss function $f_k(\bm{w})$ for each client $k$, and the global loss function $f(\bm{w}) = \sum_{k=1}^K f_k(\bm{w})$, which has minimum value $f^*$.

\begin{assumption}($\lambda$-Lipschitz Continuous Gradient)
There exists a constant $\lambda>0$, such that $\| \nabla f_k (\bm{w}) - \nabla f_k (\bm{v})\| \leq \lambda \|\bm{w}- \bm{v} \|$, $\forall \bm{w},\bm{v} \in \mathbb{R}^d$, and $k\in [K]$.
\label{assum:lipschitz}
\end{assumption}

\begin{assumption}(Bounded Global Variability)
There exists a constant $\sigma_G \geq 0$, such that the global variability of the local gradients of the cost function $\| \nabla f_k (\bm{w_t}) - \nabla f(\bm{w_t})\| \leq \sigma_G$, $\forall k \in [K]$, $\forall t$\footnote{Clients with i.i.d. label distributions may still have global variability $\sigma_G > 0$ due to having finite samples.}%although both local and global graphs are draw from the same distribution, with limited sample size of graphs $N$, the .}.
\label{assum:bounded_global}
\end{assumption}

\begin{assumption}(Bounded Gradient)
There exists a constant $G \geq 0$, such that the local gradients of the cost function $\| \nabla f_k (\bm{w_t}) \| \leq G$, $\forall k \in [K]$, $\forall t$.
\label{assum:bounded_gradient}
\end{assumption}

Assumptions~\ref{assum:lipschitz}, ~\ref{assum:bounded_global} and ~\ref{assum:bounded_gradient} are standard in the federated learning literature~\citep{li2019convergence, yu2019parallel, yang2021achieving}. 
% We then introduce another assumption to quantify the information loss:
% \begin{assumption}(Bounded Information Loss)
% There exists $\sigma_I \geq 0$, such that the information loss of the cost function is bounded by $\| \nabla \tilde{F}_k (\bm{w_t}) - \nabla f_k (\bm{w_t}) \| \leq \sigma_I$, $\forall k \in [K],t$.
% \label{assum:bounded_infoloss}
% \end{assumption}
%\subsubsection{Main Theory}
We consider a two-layer GCN, though our analysis can be extended to more layers. We work from \citet{yu2019parallel}'s convergence result for FedAvg\footnote{Our analysis applies to any federated training algorithm with bounded global variability~\citep{yang2021achieving}. } to find:

%\footnote{With full-batch training, the local variability $\sigma_L = 0$.}
\begin{theorem}(Convergence Rate for FedGCN)\label{thm:convergence}
Under the above assumptions, while training with $K$ clients, $T$ global training rounds, $E$ local updates per round, and a learning rate $\eta \leq \frac{1}{\lambda}$, we have
\begin{equation}
\frac{1}{T} \sum_{t=1}^T \mathbb{E}\left[\left\|\nabla f\left(\bm{w}_{t-1}\right)\right\|^2\right] \leq \frac{2}{\eta T}\left(f\left(\bm{w}_0\right)-f^*\right)+\frac{\lambda}{K} \eta \|I_{local} - I_{glob}\|^2 + 4 \eta^2 E^2 G^2 \lambda^2,\label{equ:converge}
\end{equation}
where $f^*$ is the minimum value of $f$, $I_{local} = K \bm{X}_k^T \bm{A}_k^T  \bm{A}_k^T  \bm{A}_k \bm{A}_k \bm{X}_k$, $I_{glob} = \bm{X}^T \bm{A}^T  \bm{A}^T  \bm{A} \bm{A} \bm{X}$. %and $B = 4 \eta^2 E^2 G^2 \lambda^2$.
\end{theorem}

\begin{table*}[h]

\fontsize{9pt}{10.8pt}\selectfont
    \centering
    \begin{tabular}{|c|c|c|}
\hline
      & Non-i.i.d.                                                                                                                                                                   & i.i.d.                                                                                                                            \\ \hline
0-hop & $  (1 - \frac{1}{K^4}) \frac{N^5}{C^5} \| B^4\| + {(1 - \frac{1}{C})}^{\frac{5}{2}} (1-p)^5$                                                                                 & $  (1 - \frac{1}{K^4}) \frac{N^5}{C^5}   \| B^4\|$                                                                                \\ \hline
1-hop & $  (1 - \frac{1}{K^4}(1 +  {\textcolor{NavyBlue}{c_\alpha p}}  +  c_\mu)^{\textcolor{Plum}{2}}) \frac{N^5}{C^5} \| B^4\| + {(1 - \frac{1}{C})}^{\frac{5}{2}} (1-p)^5$        & $  (1 - \frac{1}{K^4}(1 +  {\textcolor{NavyBlue}{c_\alpha}}  +  c_\mu)^{\textcolor{Plum}{2}}) \frac{N^5}{C^5}   \| B^4\| $        \\ \hline
2-hop & $  (1 - \frac{1}{K^4}(1 +  {\textcolor{NavyBlue}{c_\alpha p}}  +  c_\mu)^{\textcolor{ForestGreen}{6}}) \frac{N^5}{C^5} \| B^4\| + {(1 - \frac{1}{C})}^{\frac{5}{2}} (1-p)^5$ & $  (1 - \frac{1}{K^4}(1 +  {\textcolor{NavyBlue}{c_\alpha}} +  c_\mu )^{\textcolor{ForestGreen}{6}}) \frac{N^5}{C^5}   \| B^4\| $ \\ \hline
\end{tabular}
    \caption{Convergence rate bounds of FedGCN with the Stochastic Block Model and data distribution from~\cite{mcmahan2017communication} We define $c_\alpha = \frac{ (1 -\mu)\alpha N(K-1)}{CK}$ and $c_\mu = \frac{\mu \alpha N(K-1)}{K}$ for the SBM; $\alpha N$ is a constant, $c_\alpha \gg c_\mu$ and $c_\mu \simeq 0$. More hops speed up the convergence from the order of 2 to 6 (highlighted as purple and green). Communication helps more when data is more i.i.d with factor $c_\alpha p $ (highlighted as blue). Non-i.i.d. data implies a longer convergence time with factor $(1-p)^5$.}
    \label{tab:Thoery_bound_convergence_rate}
    \vspace{-0.4cm}
\end{table*}

The convergence rate is thus bounded by the difference of the information provided by local and global graphs $\|I_{local} - I_{glob}\|$, which upper bounds the global variability $\| \nabla f_k (\bm{w}) - \nabla f (\bm{w}) \|$. \emph{By 1- and 2-hop communication, the local graph $\bm{A}_k$ is closer to $\bm{A}$, resulting in faster convergence}.
% \begin{proposition}(Convergence Rates for FedGCN)\label{prop:conv}
% For generic graph,  there exist a constant $\lambda_c$ such that the difference between the local and gloabl gradients $\| \nabla f_k (\bm{w}) - \nabla f (\bm{w}) \|$ can be upper-bounded by
% \begin{equation}
%     \lambda_c \|{X}_k^T \bm{A}_k^T  \bm{A}_k^T  \bm{A}_k \bm{A}_k {X}_k - Z_{glob}\|,
% \end{equation}
% where 
% \begin{equation}
%     Z_{glob} = {X}^T {A}^T  {A}^T  {A} {A} {X} \|.
% \end{equation}
% \end{proposition}

Table \ref{tab:Thoery_bound_convergence_rate} bounds $\|I_{local} - I_{glob}\|$ for 
%the FedGCN models for generic graphs. In the table, $\dot{A_k}$ and $\ddot{A}_k$ respectively denote the adjacency matrix at client $k$ with 1- and 2-hop communication, similarly for %$\dot{f}_k$, $\ddot{f}_k$, 
%$\dot{X}_k$, $\ddot{X}_k$. In the table, we give the approximate values of these bounds for 
FedGCN trained on an SBM (stochastic block model) graph, in which we assume $N$ nodes with $C$ classes. Nodes in the same (different) class have an edge between them with probability $\alpha$ ($\mu\alpha$). Appendix~\ref{appen:SBM} 
details the full SBM.
Appendix~\ref{appen:Convegence} 
derives Table \ref{tab:Thoery_bound_convergence_rate}'s results; the SBM structure allows us to develop intuitions about FedGCN's convergence with different hops, simply by knowing the node features and graph adjacency matrix (i.e., without knowing the model). Appendix~\ref{appen:comm} derives corresponding expressions for the required communication costs. 
We validate these results in experiments and Appendix~\ref{subsec:validation}, and make the following \textbf{key observations}:
\begin{itemize}
    \item \textbf{Faster convergence with more communication hops}: 1-hop and 2-hop communication accelerate the convergence with factors $(1+c_\alpha p + c_\mu)^2$ and $(1+c_\alpha p + c_\mu)^6$, respectively. When the i.i.d control parameter $p$ increases, the difference among no (0-hop), 1-hop, and 2-hop communication increases: communication helps more when data is more i.i.d. %For a fixed number of hops, the convergence is faster as the data become more i.i.d., by a factor of $(1-p)^5$. 
    \item \textbf{More hops are needed for cross-device FL}: When the number of clients $K$ increases, as in cross-device federated learning (FL), the convergence takes longer by a factor $\frac{1}{{K}^4}$, but 2-hop communication can recover more edges in other clients to speed up the training.
    \item \textbf{One-hop is sufficient for cross-silo FL}: If the number of clients $K$ is small, approximation methods via one-hop communication can balance the convergence rate and communication overhead. We experimentally validate this intuition in the next section.
\end{itemize}

%% file: 6-experiment.tex
\section{Experimental Validation}\label{sec:experiment}
%\subsection{Dataset}
%We validate FedGCN's performance relative to previously proposed algorithms on multiple real datasets. Compared to these algorithms, 
We experimentally show that FedGCN converges to a more accurate model with less communication compared to previously proposed algorithms. We further validate Section~\ref{sec:theory}'s theoretical observations. %findings that the communication hops affect convergence less for more non-i.i.d. data.

\subsection{Experiment Settings}
We use the Cora (2708 nodes, 5429 edges), Citeseer (3327 nodes, 4732 edges), {Ogbn-ArXiv} (169343 nodes, 1166243 edges), and {Ogbn-Products} (2449029 nodes, 61859140 edges) datasets to predict document labels and Amazon product categories~\citep{wang2020microsoft,hu2020open}. %(Table \ref{tab:datasets}). %The tasks are to predict document labels and the category of Amazon products, respectively.
%Pubmed

%In Amazon product co-purchasing network, the task is to predict the category of a product. 
%In each dataset, nodes represent documents and have bag-of-words feature vectors. %for each document and a list of citation links between documents. 
%{ Citation links %between documents 
%are (undirected) edges.
%, giving a binary adjacency matrix $A$
%We aim to predict document labels.}
 %Each document has a class label, which we wish to predict. %The statistics of the datasets are summarized in Table \ref{tab:datasets}.%how many for training, how many for testing

%\subsection{Methods}
\textbf{Methods Compared.} \textbf{Centralized GCN} assumes a single client has access to the entire graph. %uses the settings in Section \ref{subsec:centralized-gcn}. 
%It has neither information loss nor communication;
{\textbf{Distributed GCN}~\citep{scardapane2020distributed} trains GCN in distributed clients, which requires communicating node features and hidden states of each layer.}
\textbf{FedGCN (0-hop)} (Section~\ref{subsec:federated-gcn}) is equivalent to federated training without communication {(\textbf{FedGraphnn})} \citep{wang2020graphfl,zheng2021asfgnn,he2021fedgraphnn}. \textbf{BDS-GCN}~\citep{wan2022bns} randomly samples cross-client edges in each global training round, while \textbf{FedSage+}~\citep{zhang2021subgraph} recovers missing neighbors by learning a linear predictor based on the node embedding, using cross-client information in each training round.
It is thus an approximation of \textbf{FedGCN (1-hop)}, which communicates the 1-hop neighbors' information across clients. \textbf{FedGCN (2-hop)} communicates the two-hop neighbors' information across clients.
%Based on the setting in Sec. \ref{subsubsec:2-order}, it communicates the 2-hop neighbor information.

We consider the Dirichlet label distribution with parameter $\beta$, as shown in Figure~\ref{fig:clients_label_distribution}. %For Cora and Citeseer, we use a two-layer GCN as in~\citet{kipf2016semi}. 
For Cora and Citeseer, we use a 2-layer GCN with~\citet{kipf2016semi}'s hyper-parameters. For Ogbn-Arxiv and Ogbn-Products, we respectively use a 3-layer GCN and a 2-layer GraphSage with \citet{hu2020open}'s hyper-parameters. We average our results over 10 experiment runs. Detailed experimental setups and extended results, including an evaluation of the HE overhead, are in Appendix~\ref{FHE_eval} and ~\ref{appen:configuration}.

%an i.i.d. data distribution, in which nodes are partitioned across clients uniformly at random, and a non-i.i.d. distribution, in which each client only contains nodes with the same (randomly chosen) label. Partial-i.i.d. settings sample a fraction of data in each manner.

\subsection{Effect of Cross-Client Communication}
We first evaluate our methods under i.i.d. $(\beta = 10000)$ and non-i.i.d. $(\beta = 100,1)$ Dirichlet data distributions on the four datasets to illustrate FedGCN's performance relative to the centralized, BDS-GCN, and FedSage+ baselines under different levels of communication. 
\begin{table}[ht]
\scriptsize{
\begin{tabular}{|l|lll|lll|ll}
\cline{1-7}
                & \multicolumn{3}{c|}{Cora, 10 clients}                                                                                                     & \multicolumn{3}{c|}{Citeseer, 10 clients}                                                                                                  &                      &                      \\ \cline{1-7}
Centralized GCN & \multicolumn{3}{c|}{0.8069$\pm$0.0065}                                                                                        & \multicolumn{3}{c|}{0.6914$\pm$0.0051}                                                                                         &                      &                      \\ \cline{1-7}
                & \multicolumn{1}{l|}{$\beta$ = 1}               & \multicolumn{1}{l|}{$\beta$ = 100}              & $\beta$ = 10000            & \multicolumn{1}{l|}{$\beta$ = 1}                & \multicolumn{1}{l|}{$\beta$ = 100}              & $\beta$ = 10000            &                      &                      \\ \cline{1-7}
FedGCN(0-hop)   & \multicolumn{1}{l|}{0.6502$\pm$0.0127}         & \multicolumn{1}{l|}{0.5958$\pm$0.0176}          & 0.5992$\pm$0.0226          & \multicolumn{1}{l|}{0.617$\pm$0.0118}           & \multicolumn{1}{l|}{0.5841$\pm$0.0168}          & 0.5841$\pm$0.0138          &                      &                      \\ \cline{1-7}
BDS-GCN         & \multicolumn{1}{l|}{0.7598$\pm$0.0143}         & \multicolumn{1}{l|}{0.7467$\pm$0.0117}          & 0.7479$\pm$0.018           & \multicolumn{1}{l|}{0.6709$\pm$0.0184}          & \multicolumn{1}{l|}{0.6596$\pm$0.0128}          & 0.6582$\pm$0.01            &                      &                      \\ \cline{1-7}
FedSage+        & \multicolumn{1}{l|}{0.8026$\pm$0.0054}         & \multicolumn{1}{l|}{0.7942$\pm$0.0075}          & 0.796$\pm$0.0075           & \multicolumn{1}{l|}{0.6977$\pm$0.0097}          & \multicolumn{1}{l|}{0.6856$\pm$0.0121}          & 0.688$\pm$0.0086           &                      &                      \\ \cline{1-7}
FedGCN(1-hop)   & \multicolumn{1}{l|}{\textbf{0.81$\pm$0.0066}}  & \multicolumn{1}{l|}{0.8009$\pm$0.007}           & 0.8009$\pm$0.0077          & \multicolumn{1}{l|}{\textbf{0.7006$\pm$0.0071}} & \multicolumn{1}{l|}{0.6891$\pm$0.0067}          & 0.693$\pm$0.0069           &                      &                      \\ \cline{1-7}
FedGCN(2-hop)   & \multicolumn{1}{l|}{0.8064$\pm$0.0043}         & \multicolumn{1}{l|}{\textbf{0.8084$\pm$0.0051}} & \textbf{0.8087$\pm$0.0061} & \multicolumn{1}{l|}{0.6933$\pm$0.0067}          & \multicolumn{1}{l|}{\textbf{0.6953$\pm$0.0069}} & \textbf{0.6948$\pm$0.0032} &                      &                      \\ \cline{1-7}
                & \multicolumn{3}{c|}{Ogbn-Arxiv, 10 clients}                                                                                               & \multicolumn{3}{c|}{Ogbn-Products, 5 clients}                                                                                             & \multicolumn{1}{c}{} & \multicolumn{1}{c}{} \\ \cline{1-7}
Centralized GCN & \multicolumn{3}{c|}{0.7$\pm$0.0082}                                                                                           & \multicolumn{3}{c|}{0.7058$\pm$0.0008}                                                                                         & \multicolumn{1}{c}{} & \multicolumn{1}{c}{} \\ \cline{1-7}
                & \multicolumn{1}{l|}{$\beta$ = 1}               & \multicolumn{1}{l|}{$\beta$ = 100}              & $\beta$ = 10000            & \multicolumn{1}{l|}{$\beta$ = 1}                & \multicolumn{1}{l|}{$\beta$ = 100}              & $\beta$ = 10000            &                      &                      \\ \cline{1-7}
FedGCN(0-hop)   & \multicolumn{1}{l|}{0.5981$\pm$0.0094}         & \multicolumn{1}{l|}{0.5809$\pm$0.0017}          & 0.5804$\pm$0.0015          & \multicolumn{1}{l|}{0.6789$\pm$0.0031}          & \multicolumn{1}{l|}{0.658$\pm$0.0008}           & 0.658$\pm$0.0008           &                      &                      \\ \cline{1-7}
BDS-GCN         & \multicolumn{1}{l|}{0.6769$\pm$0.0086}         & \multicolumn{1}{l|}{0.6689$\pm$0.0024}          & 0.6688$\pm$0.0015          & \multicolumn{1}{l|}{0.6996$\pm$0.0019}          & \multicolumn{1}{l|}{0.6952$\pm$0.0012}          & 0.6952$\pm$0.0009          &                      &                      \\ \cline{1-7}
FedSage+        & \multicolumn{1}{l|}{0.7053$\pm$0.0073}         & \multicolumn{1}{l|}{0.6921$\pm$0.0014}          & 0.6918$\pm$0.0024          & \multicolumn{1}{l|}{0.7044$\pm$0.0017}          & \multicolumn{1}{l|}{0.7047$\pm$0.0009}          & 0.7051$\pm$0.0006          &                      &                      \\ \cline{1-7}
FedGCN(1-hop)   & \multicolumn{1}{l|}{0.7101$\pm$0.0078}         & \multicolumn{1}{l|}{\textbf{0.6989$\pm$0.0038}} & 0.7004$\pm$0.0031          & \multicolumn{1}{l|}{0.7049$\pm$0.0016}          & \multicolumn{1}{l|}{\textbf{0.7057$\pm$0.0003}} & \textbf{0.7057$\pm$0.0004} &                      &                      \\ \cline{1-7}
FedGCN(2-hop)   & \multicolumn{1}{l|}{\textbf{0.712$\pm$0.0075}} & \multicolumn{1}{l|}{0.6972$\pm$0.0075}          & \textbf{0.7017$\pm$0.0081} & \multicolumn{1}{l|}{\textbf{0.7053$\pm$0.002}}  & \multicolumn{1}{l|}{\textbf{0.7057$\pm$0.0009}} & 0.7055$\pm$0.0006          &                      &                      \\ \cline{1-7}
\end{tabular}
}
\caption{Test Accuracy on four datasets, for i.i.d.$(\beta = 10000)$ and non-i.i.d. $(\beta = 100,1)$ data. FedGCN (1-hop,2-hop) performs best on i.i.d. and non-i.i.d. data, and FedGCN (0-hop) has the most information loss. We assume FedSage+'s linear approximator perfectly learns neighbor information.}
\vspace{-0.5cm}
\label{tab:accuracy_comparison}
\end{table}

As shown in Table \ref{tab:accuracy_comparison}, \emph{FedGCN(1-, 2-hop) has higher test accuracy than FedSage+ and BDS-GCN}, reaching the same test accuracy as centralized training in all settings.  FedGCN(1-, 2-hop) is able to converge faster to reach the same accuracy with the optimal number of hops depending on the data distribution. In such a cross-silo setting (10 clients), FedGCN(1-hop) achieves a good tradeoff between communication and model accuracy. \emph{FedGCN (0-hop) performs worst in the i.i.d. and non-i.i.d. settings, due to information loss from cross-client edges}.

\textbf{Why Non-i.i.d Has Better Accuracy in 0-hop?} In Table~\ref{tab:accuracy_comparison} with 10 clients, 0-hop has better accuracy since non-i.i.d. has fewer cross-client edges (less information loss) than i.i.d as in theoretical analysis.

\begin{figure*}[ht]
    \includegraphics[width = 0.32\textwidth]{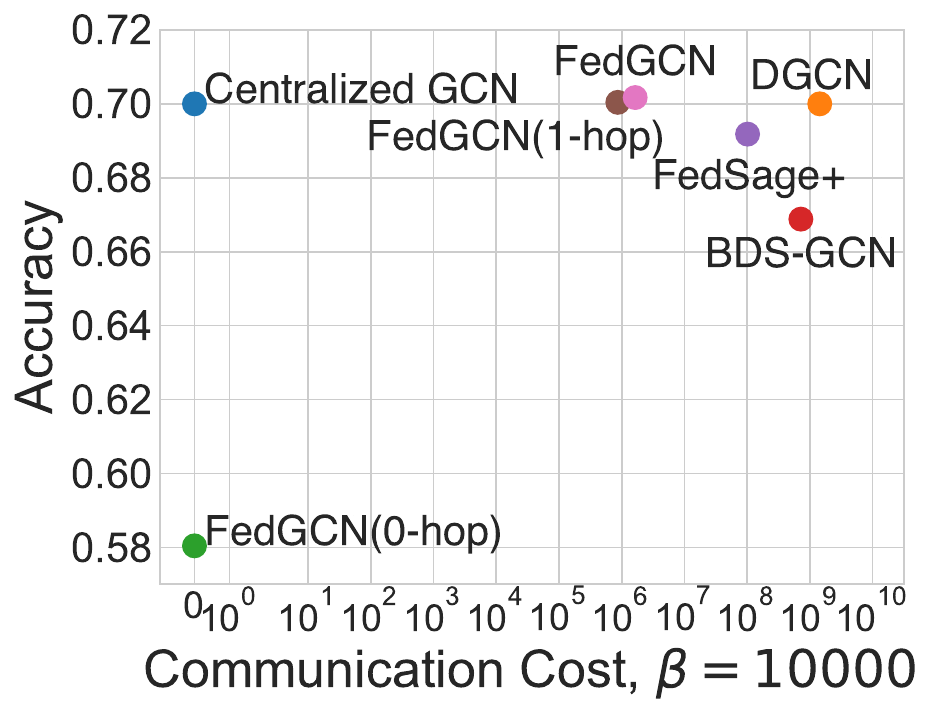}
    \includegraphics[width = 0.32\textwidth]{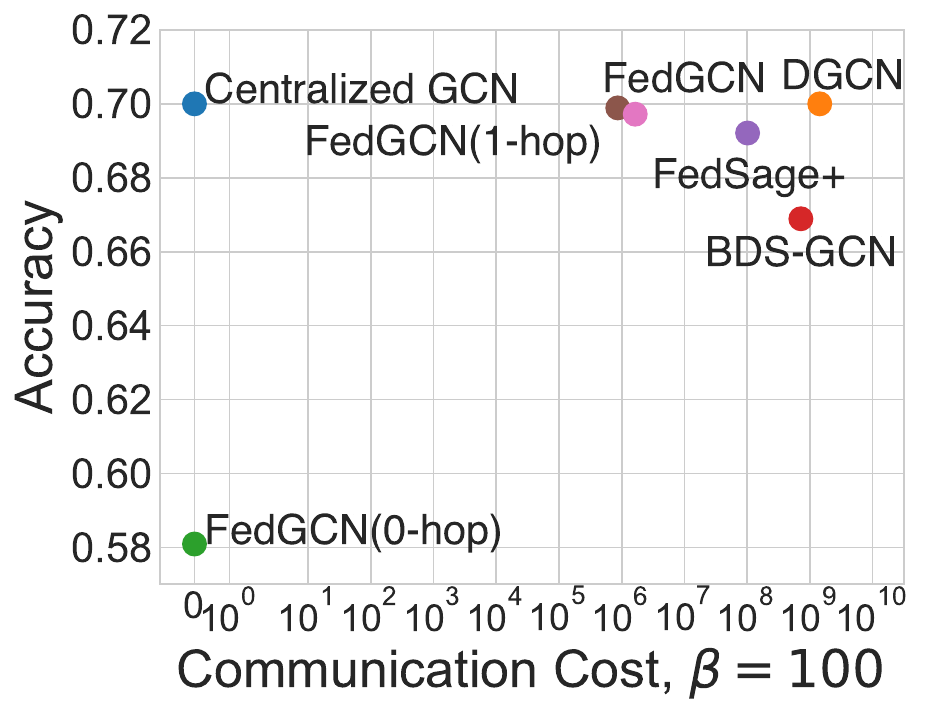}
    \includegraphics[width = 0.32\textwidth]{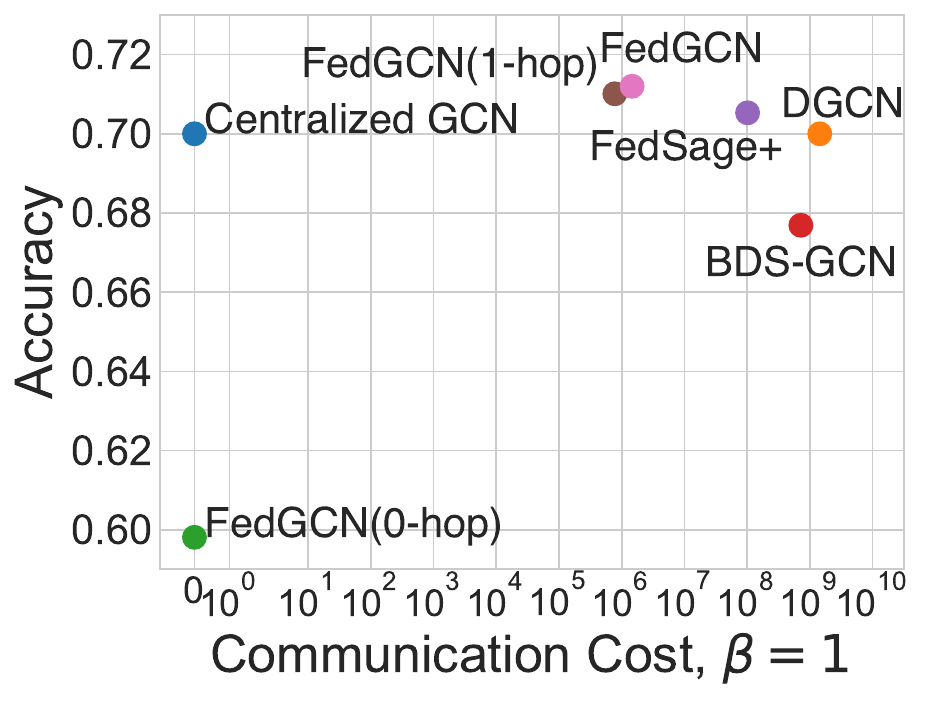}
    \caption{{Test accuracy vs. communication cost until convergence of algorithms in the i.i.d., partial-i.i.d. and non-i.i.d. settings for the OGBN-ArXiv dataset. FedGCN uses orders of magnitude less communication (at least 100$\times$) than BDS-GCN and FedSage+, while achieving higher test accuracy.}}
    \label{fig:communication_with_accuracy}
    \vspace{-0.2cm}
\end{figure*}

\textbf{Communication Cost vs. Accuracy.} Figure~\ref{fig:communication_with_accuracy} shows the communication cost and test accuracy of different methods on the OGBN-ArXiv dataset. \emph{FedGCN (0-, 1-, and 2-hop) requires little communication with high accuracy}, while Distributed GCN, BDS-GCN and {FedSage+} require communication at every round, incurring over 100$\times$ the communication cost of any of FedGCN's variants. FedGCN (0-hop) requires much less communication than 1- and 2-hop FedGCN, 
but has lower accuracy due to information loss, displaying a convergence-communication tradeoff. Both 1- and 2-hop FedGCN achieve similar accuracy as centralized GCN, indicating that a 1-hop approximation of cross-client edges is sufficient in practice to achieve an accurate model.

\begin{figure*}[ht]
    \centering
    \includegraphics[width = 0.32\textwidth]{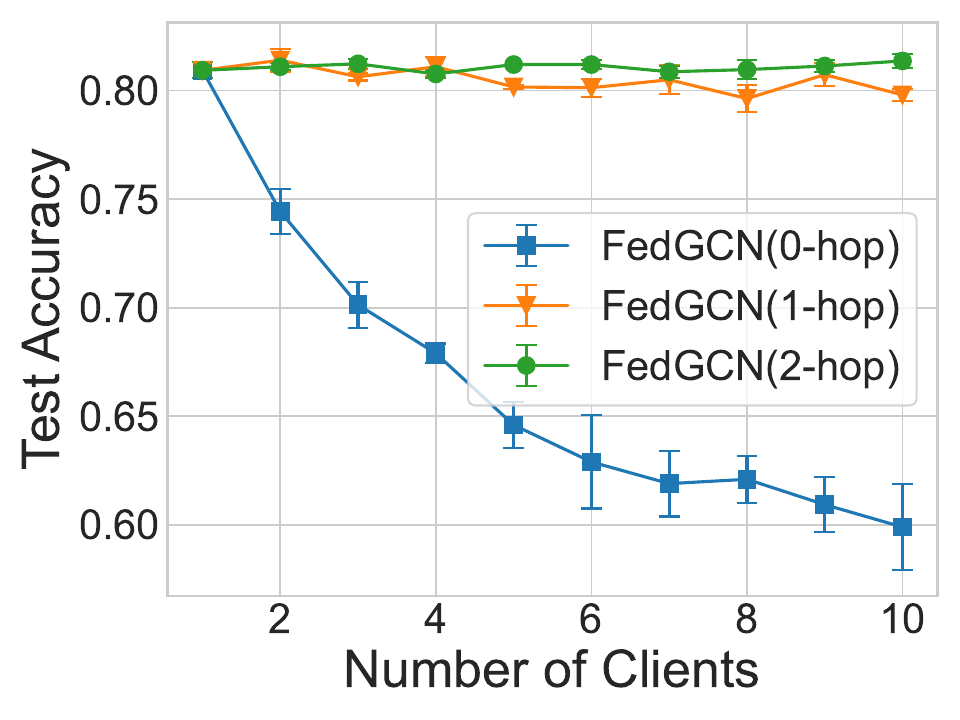}
    \includegraphics[width = 0.32\textwidth]{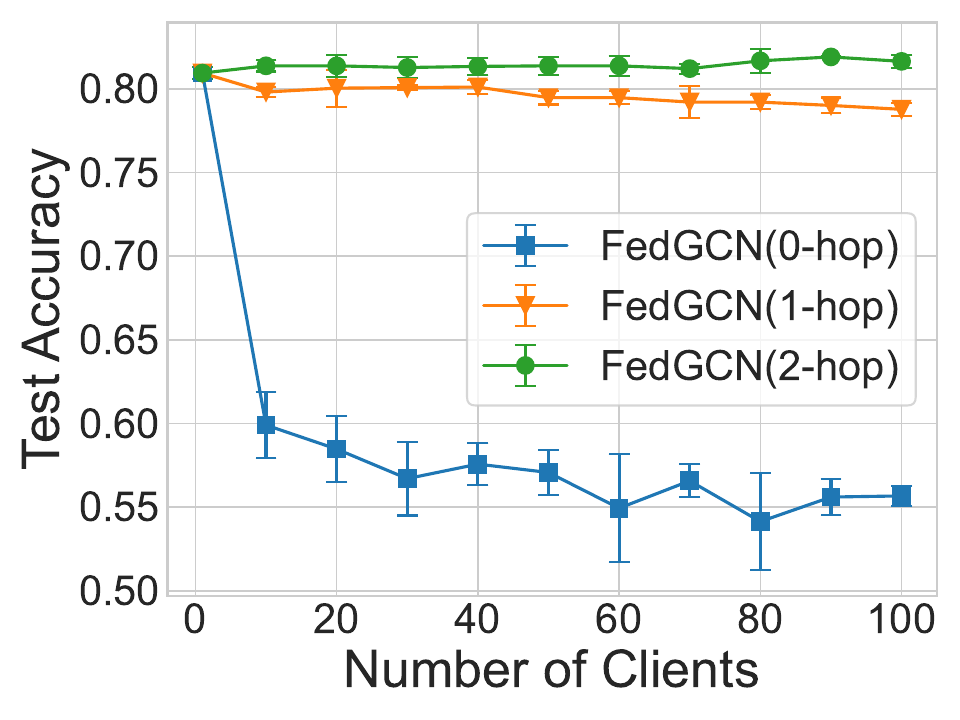}
    \includegraphics[width = 0.32\textwidth]{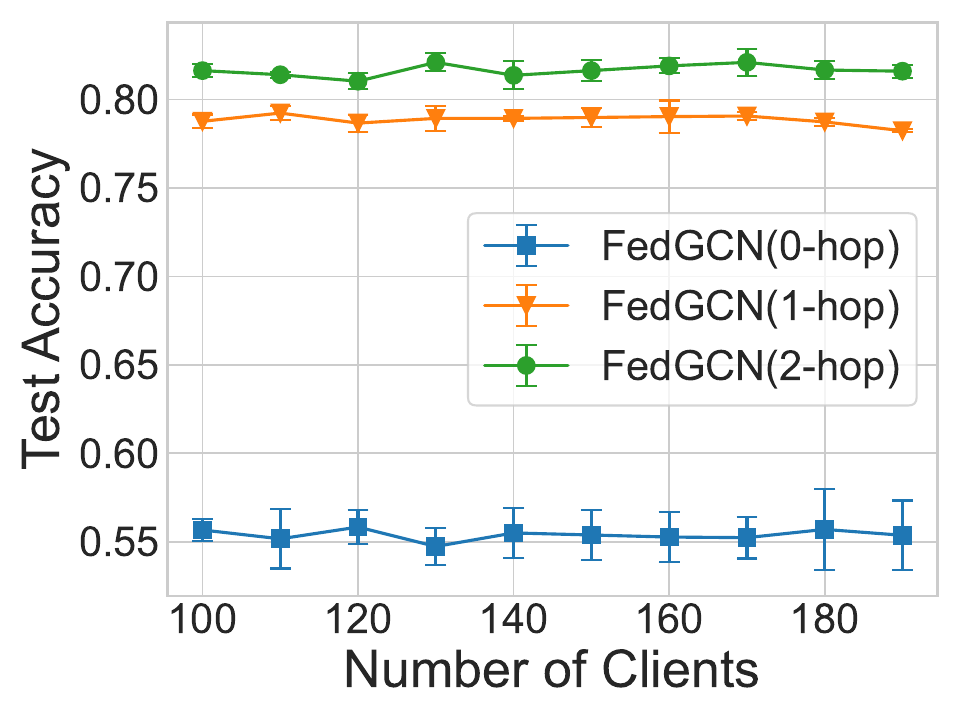}
    \caption{Test accuracy with the number of clients on the Cora dataset.}
    \label{fig:cross_device_cross_silo}
    \vspace{-0.5cm}
\end{figure*}

\textbf{Cross-Silo and Cross-Device Training.} 
As shown in Figure~\ref{fig:cross_device_cross_silo} (left), 1-hop and 2-hop communication achieve similar test accuracy in the cross-silo setting with few clients, though 1-hop communication has lower communication costs. However, in the cross-device setting with many clients (Figure~\ref{fig:cross_device_cross_silo}~right), the 1-hop test accuracy drops with more clients, indicating that 2-hop communication may be necessary to maintain high model accuracy, as suggested by our theoretical analysis.

%% file: 7-conclusion.tex
\section{Conclusion}\label{sec:conclusion}
We propose FedGCN, a framework for federated training of graph convolutional networks for semi-supervised node classification. The FedGCN training algorithm is based on the insight that, although distributed GCN training typically ignores cross-client edges, these edges often contain information useful to the model, which can be sent in a single round of communication before training begins. FedGCN allows for different levels of communication to accommodate different privacy and overhead concerns, with more communication generally implying less information loss and faster convergence, and further integrates HE for privacy protection. We quantify FedGCN's convergence under different levels of communication and different degrees of non-i.i.d. data across clients and show that FedGCN achieves high accuracy on real datasets, with orders of magnitude less communication than previous algorithms.

%% file: 8-use_case.tex
\section{Applications of FedGCN}
In this section, we discuss three important yet challenging applications of federated graph learning. Compared with prior works, our FedGCN can overcome the challenges of each application setting without sacrificing accuracy.

\subsection{Multi-step Distributed Training with $1000\times$ Less Communication Cost}
In distributed training, the main focus is to train a model with fast computation time and high accuracy, utilizing the resources of multiple computing servers. Privacy is not a concern in this case. FedGCN requires much less communication cost compared with distributed training methods (e.g., BDS-GCN~\cite{wan2022bns}) since FedGCN only requires pre-training communication. Moreover, FedGCN suggests that non-i.i.d. data distributions can further reduce communication, since non-i.i.d. partition results in fewer cross-client edges. FedGCN can first partition the graph to non-i.i.d. and perform precomputation to minimize the communication cost while maintaining model accuracy.

\subsection{A Few Clients with Large Graphs in Federated Learning}
Due to privacy regulations (e.g., GDPR in Europe) across countries, some data, e.g., that collected by Internet services like social networks, needs to be localized in each country or region. Here, each country represents a client, and ``nodes'' might be citizens of that country who use a particular service, g., users of a social network. Users in different countries (at different clients) may then interact, creating a cross-client edge. In this case, a single large country might be able to train models with sufficient performance using only its users' information, but some small countries might find it hard to train a good model as they have fewer constituent nodes. Federated learning for training a global model across countries, i.e., cross-silo federated learning, can help to train models in this setting, and there are relatively few cross-client edges compared to in-client edges. For example, in social networks, edges represent the connections between users, and users are more closely connected within a country. However, the small amount of cross-country (cross-client) edges might seriously affect the model performance since these edges can be more important for decision-making than the in-client edges. For example, for anomaly detection on payment records, cross-country transactions are the key to detecting international money laundering and fraud, and ignoring these edges makes it impossible to detect these behaviors. FedGCN can take these edges into account and the trainers can decide if they want these edges based on the edge utility for their tasks.

FedGCN only communicates accumulated and homomorphically encrypted neighbor information at the initial round with better privacy guarantees and can also add differential privacy to better fit privacy regulations.

\subsection{Millions of Clients with Small Graphs in Federated Learning}

With the development of IoT (Internet-of-Things) devices, many people own several devices that can collect, process, and communicate data (e.g., Mobile phones, smart watches, or computers). Recently, smart home devices (e.g. cameras, light bulbs, and smart speakers) have also been adopted by millions of users, e.g., to monitor home security, the health care of senior citizens and infants, and package delivery. %It is changing people's lives with more and more applications after the pandemic. 
These mobile devices, and the applications that run on them, can also have interactions and connect with these IoT devices, e.g., unlocking the front door triggers the living room camera. %\carlee{give example of an interaction?}

All these devices and applications are connected and form a graph, in which the devices/applications are nodes and their interactions are edges. The information of local devices or applications (node features) can then be very privacy sensitive, e.g., it may include video recording, accurate user location, and other information about users' personal habits. Federated training keeps the data localized to maintain privacy, while cross-client edges affect the federated training performance with privacy leakage. Here we take a ``client'' to mean a single user, who may own several devices or applications (i.e., nodes in the graph). Millions of devices across multiple users can then be connected, although each user (client) owns only a few devices, which means there are many edges across clients. The huge amount of cross-client edges can then seriously affect federated training's performance. The condition becomes more serious with the co-optimization of heterogeneous data distribution with limited local data.

Our FedGCN can significantly improve both the convergence time and model accuracy since it does not have information loss regardless of the number of clients. Since the number of cross-client edges increases with the number of clients, prior methods that ignore these edges or communicate (some of) their information in every round respectively face more information loss and communication costs.

%% file: 9-future_directions_and_applications.tex
\section{Future Directions}

Although the FedGCN can overcome the challenges mentioned above, it mainly works on training accumulation-based models like GCN and GraphSage. \emph{Our first theoretical analysis on federated graph learning with cross-client edges can be further improved}. Several open problems in federated graph learning also need to be explored.

\textbf{Federated Training of Attention-based GNNs}
Attention-based GNNs like GAT (Graph attention network) require calculating the attention weights of edges during neighbor feature aggregation, where the attention weights are based on the node features on both sides of edges and attention parameters. The attention parameters are updated at every training iteration and cannot be simply fixed at the initial round. How to train attention-based GNNs in a federated way with high performance and privacy guarantees is an open challenge and promising direction.

\textbf{Neighbor Node and Feature Selection to Optimize System Performance} %Some graph edges reside across devices. 
General federated graph learning optimizes the system by only sharing local models, without utilizing cross-device graph edge information,  which leads to less accurate global models. On the other hand, communicating massive additional graph data among devices introduces communication overhead and potential privacy leakage. To save the communication cost without affecting the model performance, one can select key neighbors and neighbor features to reduce communication costs and remove redundant information. For privacy guarantee, if there is one neighbor node, it can be simply dropped to avoid private data communication. FedGCN can be extended by using selective communication in its pre-training communication round.

\textbf{Integration with $L$-hop Linear GNN Approximation methods}
To speed up the local computation speed, $L$-hop Linear GNN Approximation methods use precomputation to reduce the training computations by running a simplified GCN ($\bm{A}^L \bm{X} \bm{W}$ in SGC~\cite{wu2019simplifying}, $[\bm{A}\bm{X}\bm{W}, \bm{A}^2 \bm{X}\bm{W}, …, \bm{A}^L \bm{X}\bm{W}]$ in SIGN~\cite{frasca2020sign}, and $\bm{\Pi} \bm{X} \bm{W}$ in PPRGo~\cite{bojchevski2020scaling} where $\bm{\Pi}$ is the pre-computed personalized pagerank), but the communication cost is not reduced if we perform these methods alone. They are thus a complementary approach for efficient GNN training. FedGCN (2-hop, 1-hop) changes the model input ($\bm{A}$ and $\bm{X}$) to reduce communication in the FL setting, but the GCN model itself is not simplified. FedGCN can incorporate these methods to speed up the local computation, especially in constrained edge devices.

% \subsection{Federated Training of Dynamic Heterogeneous Graphs}
% Real-world graphs are often dynamic with heterogeneous node and edge types. Heterogeneous graph transformers have been proposed to deal with these graphs and show great performance for various applications. How to train the graph transformers in a federated way without private leakage becomes even more challenging. 

% \subsection{Federated Multi-Task Graph Training} Different devices may run different related graph learning tasks (e.g. node classification, link prediction, and graph classification). Separate federated training for each task, despite their being related due to learning on similar graph structures, causes high overhead on each device.  Multi-task optimization also poses issues from system orchestration (e.g. how to orchestrate devices with the same task within the same multi-task FL process to reduce inter-party communication).

% \subsection{Privacy \& Security}
% With the privacy issue from cross-client graph edges,  it is critical to design a secure graph-sharing mechanism to mitigate privacy leakage without adding much cryptographic computation and communication overhead. Additionally, the privacy of current FL frameworks relies on the naive guarantee from model aggregation functions, which is subject to various attacks on both user privacy and system security if local models are shared in a plaintext view. For example, a sensor can be compromised to infer other sensors’ data.

%% file: appendix.tex
\newpage
\appendix
\onecolumn
\section{FedGCN Training Algorithm}\label{appen:algorithm}

\begin{algorithm}%[H]
\DontPrintSemicolon
\caption{\texttt{FedGCN} Federated Training for Graph Convolutional Network}
\SetNoFillComment
\label{Alg:fedgcn}
\SetKwFor{ForPar}{for}{do in parallel}{end forpar}
% \textbf{Inputs:} $C$: communication rounds, $E_i$ number of local updates in client $i$, $\gamma$ and $\eta$: local and global learning rates,  $\bm{\theta}^{(0)}$, $\bm{\phi}^{(0)}$ initial global models for GNN and Prediction, initial gradient tracking $\bm{\delta}_j^{(0)}=\bm{0}, \; \forall j\in[m]$\\[5pt]
% \tcp{Initialization Round}
% All participants agree on certain cryptocontexts and keys
\tcp{Pre-Training Communication Round}
%\tcp{Step 1}
%round by round communication
\ForPar{each client $k\in[K]$}{

            %\tcp{Gather Information} 
    %\tcp{$[\![\cdot]\!]$:Encryption}
    Send $[\![\{\sum_{j\in \mathcal{N}_i} \mathbb{I}_k (c(j))\cdot \bm{A}_{ij}  \bm{x}_j\}_{i \in {\mathcal{V}_k}}]\!]$ to the server
    }
\tcp{Server Operation}
\ForPar{ $i \in \mathcal{V}$}{
            %\tcp{Gather Information} 
    $[\![\sum_{j\in \mathcal{N}_i} \bm{A}_{ij}  \bm{x}_j ]\!] = \sum_{d=1}^C [\![ \sum_{j\in \mathcal{N}_i} \mathbb{I}_k (c(j))\cdot \bm{A}_{ij}  \bm{x}_j]\!]$
    }
%\tcp{Step 2}
\ForPar{each client $k\in[K]$}{
    \If{ 1-hop}{
    Receive 
    $[\![\{\sum_{j\in \mathcal{N}_i} \bm{A}_{ij}  \bm{x}_j\}_{i\in {\mathcal{V}_k}}]\!]$ and decrypt it
    }
    \If{ 2-hop}{
            Receive 
    $[\![\{\sum_{j\in \mathcal{N}_i} \bm{A}_{ij}  \bm{x}_j\}_{i\in \mathcal{N}_{\mathcal{V}_k}}]\!]$ and decrypt it
        }
    }

\tcp{Training Rounds}
\For{$t=1, \ldots, T$}{
    \ForPar{each client $k\in[K]$}{
             Receive 
    $[\![ {\boldsymbol{w}}^{(t)} ]\!]$ and decrypt it\\
         Set $\bm{w}_k^{(t,0)}={\boldsymbol{w}}^{(t)}$, \\
        \For{ $e=1,\ldots,E$}{
            Set $\bm{g}_{\bm{w}_k}^{(t,e)} = \nabla_{\bm{w}_k} f_k (\bm{w}_k^{(t,e-1)};\mathcal{G}_k)$\\
             $\bm{w}^{(t,e)}_{k}=\bm{w}^{(t,e-1)}_k-\eta~ {\boldsymbol{g}}^{(t,e)}_{\bm{w}_k}$
             \tcp{Update Parameters}
            %  \tcp{Update Parameters}
            %  $\bm{w}^{(t,e+1)}_{k}=\bm{w}^{(t,e)}_k-\eta_{L}~ \tilde{\boldsymbol{g}}^{(t,e)}_{\bm{w}_k}$\\
        }
        %$\bm{\Delta}_{\bm{w}_k}^{(t,E)} = \bm{w}^{(t,E)}_k - \bm{w}^{(t,0)}_k$\\
        Send $[\bm{w}^{(t,E)}_k]\!]$ to the server
    }
    
    \tcp{Server Operations}
    %~~~\tcp{Difference Aggregation}
   $[\![\bm{w}^{(t+1)}]\!]=\frac{1}{K}\sum_{d=1}^C [\![\bm{w}^{(t,E)}_k]\!]$ \tcp{Update Global Models} %need to better
    %~~~\tcp{Update Global Models} 
    \scalebox{0.85}Broadcast $[\![\bm{w}^{(t+1)}]\!]$ to local clients 
}
\end{algorithm}

\section{Background and Preliminaries}
\subsection{Federated Learning}
Federated learning was first proposed by~\cite{mcmahan2017communication}, who build decentralized machine learning models while keeping personal data on clients. Instead of uploading data to the server for centralized training, clients process their local data and occasionally share model updates with the server. Weights from a large population of clients are aggregated by the server and combined to create an improved global model. 

The FedAvg algorithm~\cite{mcmahan2017communication} is used on the server to combine client updates and produce a new global model. At training round $t$, a global model $\bm{w}^{(t)}$ is sent to $K$ client devices.

At each local iteration $e$, every client $k$ computes the gradient, $\bm{g}_{\bm{w}_k}^{(t,e)}$, on its local data by using the current model $\bm{w}_k^{(t,e-1)}$. For a client learning rate $\eta$, the local client update at the $e$-th local iteration, $\bm{w}_k^{(t,e)}$, is given by
\begin{equation}
   \bm{w}_k^{(t,e)} \leftarrow \bm{w}_k^{(t,e-1)}- \eta \bm{g}_{\bm{w}_k}^{(t,e)}.
\end{equation}
After $E$ local iterations, the server then does an aggregation of clients' local models to obtain a new global model, 

\begin{equation}
    \bm{w}^{(t+1)}= \frac{1}{K} \sum_{d=1}^C \bm{w}^{(t, E)}_k.
\end{equation}
The process then advances to the next training round, $t + 1$.

%B batch size of each client

\subsection{Graph Convolutional Network}
A multi-layer Graph Convolutional Network (GCN)~\citep{kipf2016semi} has the layer-wise propagation rule
\begin{equation}\label{equ:gcn}
    \bm{H}^{(l+1)}=\phi(\bm{A}\bm{H}^{(l)}\bm{W}^{(l)}).
\end{equation}
The weight adjacency matrix $\bm{A}$ can be normalized or non-normalized given the original graph, and $\bm{W}^{(l)}$ is a layer-specific trainable weight matrix. %can be normalized as $\widetilde{\bm{D}}^{-\frac{1}{2}}\widetilde{\bm{A}}\widetilde{\bm{D}}^{-\frac{1}{2}}$, where $\widetilde{\bm{A}}=\bm{A}+\bm{I}_N$, $\bm{I}_N$ is the identity matrix, $\widetilde{\bm{D}}_{ii}=\sum_j \widetilde{\bm{A}}_{ij}$. 
The activation function is $\phi$, typically ReLU (rectified linear units), with a softmax in the last layer for node classification. The node embedding matrix in the $l$-th layer is $\bm{H}^{(l)}\in \mathbb{R}^{N\times d}$, which contains high-level representations of the graph nodes transformed from the initial features; $\bm{H}^{(0)}=\bm{X}$.

In general, for a GCN with $L$ layers of the form \ref{equ:gcn}, the output for node $i$ will depend on neighbors up to $L$ steps away. We denote this set by $\mathcal{N}_i^L$ as $L$-hop neighbors of $i$. Based on this idea, the clients can first communicate the information of nodes. After the communication of information, we can then train the model.

\subsection{Stochastic Block Model}\label{appen:SBM} For positive integers $C$ and $N$, a probability vector $\bm{p}\in [0,1]^C$, and a symmetric connectivity matrix $\bm{B}\in[0,1]^{C\times C}$, the SBM defines a random graph with $N$ nodes split into $C$ classes. The goal of a prediction method for the SBM is to correctly divide nodes into their corresponding classes, based on the graph structure. Each node is independently and randomly assigned a class in $\{1,...,C\}$ according to the distribution $\bm{p}$; we can then say that a node is a ``member'' of this class. Undirected edges are independently created between any pair of nodes in classes $c$ and $d$ with probability $\bm{B}_{cd}$,
%Each node only belongs to one class. Not sure the former description is clear enough 
where the $(c,d)$ entry of $\bm{B}$ is 
\begin{equation}
\bm{B}_{cd}=\left\{
\begin{aligned}
 \alpha& ,\;c=d\\
\mu \alpha & ,\;c\neq d,
\end{aligned}
\right.
\end{equation}
for $\alpha\in(0,1)$ and $\mu\in(0,1)$, implying that the probability of an edge forming between nodes in the same class is $\alpha$ (which is the same for each class) and the edge formation probability between nodes in different classes is $\mu \alpha$.

Let $\bm{Y} \in {\{0,1\}}^{N\times C}$ denotes the matrix representing the nodes' class memberships, where $\bm{Y}_{ic}=1$ indicates that node $i$ belongs to the $c$-th class, and is $0$ otherwise. 
We use $\bm{A}\in\{0,1\}^{N \times N}$ to denote the (symmetric) adjacency matrix of the graph, where $\bm{A}_{ij}$ indicates whether there is a connection (edge) between node $i$ and node $j$. From our node connectivity model, we find that given $\bm{Y}$, for $i<j$, we have
\begin{equation}
    \bm{A}_{ij}|\{\bm{Y}_{ic}=1,\bm{Y}_{jd}=1\} \backsim \text{Ber}(\bm{B}_{cd}),
\end{equation} 
where $\text{Ber}(p)$ indicates a Bernoulli random variable with parameter $p$. Since all edges are undirected, $\bm{A}_{ij}=\bm{A}_{ji}$. We further define the connection probability matrix $\bm{P}=\bm{Y} \bm{B} \bm{Y}^T \in [0,1]^{N\times N}$, where $\bm{P}_{ij}$ is the connection probability of node $i$ and node $j$ and
%\begin{equation}
    $\mathbb{E}[\bm{A}]=\bm{P}$. %-\text{diag}(P)
%\end{equation}

\section{Training Configuration}\label{appen:configuration}

\subsection{Statistics of Datasets}
\begin{table}[ht]
    \centering
    \begin{tabular}{c|c|c|c|c}
    \hline
         Dataset& Nodes &Edges & Features &Classes \\
    \hline
          Cora & 2,708 & 5,429 & 1,433 &7 \\
          Citeseer & 3,327 & 4,732 & 3,703 &6 \\
          %Pubmed & 19,717 & 44,338 & 500 &3 \\
          Ogbn-Arxiv & 169,343 & 1,166,243 & 128 &40 \\
          Ogbn-Products & 2,449,029 & 61,859,140 & 100 &47 \\
    \hline
    \end{tabular}
    \caption{Statistics of datasets.}
    \label{tab:datasets}
\end{table}

\subsection{Experiment Hyperparameters}

For Cora and Citeseer, we use a two-layer GCN with ReLU activation for the first and Softmax for the second layer, as in~\citet{kipf2016semi}. There are 16 hidden units. A dropout layer between the two GCN layers has a dropout rate of 0.5. We use 300 training rounds with the SGD optimizer for all settings with a learning rate of 0.5, L2 regularization $5 \times 10^4$, and 3 local steps per round for federated settings. For the OGBN-Arxiv dataset, we instead use a 3-layer GCN with 256 hidden units and 600 training rounds. For the OGBN-Products dataset, we use 2-layer GraphSage, 256 hidden units, and 450 training rounds. All settings are the same as the papers~\cite{kipf2016semi, hu2020open}.
The local adjacency matrix is normalized by $\tilde{\bm{A}} =\bm{D}^{-\frac{1}{2}}\bm{A}\bm{D}^{-\frac{1}{2}}$ when using GCN. We evaluate the local test accuracy given the local graph $\mathcal{G}_k$ and get the average test accuracy of all clients as the global test accuracy. We set the number of clients to 10 and averaged over 10 experiment runs. %for better evaluation of the non-i.i.d. condition. We show averages of 10 experiment runs. 

\subsection{Computation Resource}
Experiments are done in a p3d.16xlarge instance with 8 GPUs (32GB memory for each GPU) and 10 g4dn.xlarge instances (16GB GPU memory in each instance). One run of the OGBN-Products experiment can take 20 minutes due to full-batch graph training.

\section{Communication Cost and Tradeoffs}

%In the SBM, nodes in the same (different) class have an edge between them with probability $\alpha$ ($\mu\alpha$), $\mu \in[0, 1]$. Appendix A %\ref{appen:SBM} 
%to analysis the affect of data distribution. Link probability between nodes within the same class is $\alpha$ while nodes between different classes have lower link probability $\mu \alpha$. 
In this appendix, we examine the communication cost, and the resulting convergence-communication tradeoff, of FedGCN. As for the convergence analysis in Section~\ref{sec:theory}, we derive communication costs for general graphs and then more interpretable results for the SBM model.

\begin{proposition}(Communication Cost for FedGCN)\label{prop:comm}
For $L$-hop communication of GCNs with a number of layers $\geq L$, the size of messages from $K$ clients to the server in a generic graph is 
\begin{equation}
    \sum_{i\in \mathcal{V}} |c(\mathcal{N}_i) | d + \sum_{k=1}^{K} |\mathcal{N}_k^{L-1}| d,
\end{equation} 
where $c(\mathcal{N}_i)$ denotes the set of clients storing the neighbors of node $i$.

% For data distribution with a factor $\bm{p}$ on SBM model, the expected size of the first part, $\sum_{i\in \mathcal{V}} |c(\mathcal{N}_i) | d $, is 
% \begin{equation}
%     N(1 + (C-1)(1 - (1-\alpha)^{\frac{Np}{C^2}}(1-\mu \alpha)^{\frac{N(C-p)}{C^2}}))d. 
% \end{equation}

\begin{table*}[ht]
\fontsize{8.6pt}{10.32pt}\selectfont
    \centering
    \begin{tabular}{c|c|c|c}
    \hline
        %SBM 
        Data Distribution & 0-hop & 1-hop  & 2-hop\\
    \hline
    
        Generic Graph & 0 & $\sum_{i\in \mathcal{V}} |c(\mathcal{N}_i) | d + N d$ 
        & $\sum_{i\in \mathcal{V}} |c(\mathcal{N}_i) | d + \sum_{k=1}^{K} |\mathcal{N}_{\mathcal{V}_k}| d$\\
        Non-i.i.d. (SBM) & 0 &  $ ( c_\mu + 2)Nd $  & $ 2( c_\mu + 1)Nd $\\
        Partial-i.i.d. (SBM) & 0 & $ (c_\alpha p  +  c_\mu + 2)Nd $ & $ 2(c_\alpha p  +  c_\mu + 1)Nd $\\
        i.i.d. (SBM) & 0 &  $ (c_\alpha  +  c_\mu + 2)Nd $  & $ 2(c_\alpha  +  c_\mu + 1)Nd $ \\
    \hline
    \end{tabular}
    \caption{Communication costs. $|.|$ denotes the size of the set and $\sum_{i\in \mathcal{V}} |c(\mathcal{N}_i) | d $ is the cost of the message that the server received from all clients, where $c(\mathcal{N}_i)$ denotes the set of clients storing the neighbors of node $i$. Communication cost increases with the i.i.d control parameter $\bm{p}$. 2-hop communication has around twice the cost of 1-hop communication.}
    \label{tab:SBM_communication_cost}
\end{table*}

For a better understanding of the above form, Table \ref{tab:SBM_communication_cost} gives the approximated (assuming $\alpha,\mu << 1$) size of messages between clients for i.i.d. and non-i.i.d. data, for generic graphs and an SBM with $N$ nodes and $d$-dimensional node features. Half the partial i.i.d. nodes are chosen in the i.i.d. and half the non-i.i.d. settings.
\end{proposition}
Appendix \ref{appen:comm} 
proves this result. %The 1-hop communication requires to communicate with the neighbors of nodes stored in other clients. 
In the non-i.i.d. setting, most nodes with the same labels are stored in the same client, which means there are much fewer edges linked to nodes in the other clients than in the i.i.d. setting, incurring much less communication cost (specifically, $c_\alpha Nd$ fewer communications) for 1- and 2-hop FedGCN. Note that communication costs vary with $N$ but not $K$, the number of clients, as clients communicate directly with the server and not with each other.

% \begin{table*}[h]
%     \centering
%     \begin{tabular}{c|c|c|c}
%     \hline
%         Convergence Rate & 0-hop Local vs Global & 1-hop Local vs Global & 2-hop Local vs Global\\
%     \hline
%         IID & & & \\
%         Non-IID & & & \\
%     \hline
%     \end{tabular}
%     \caption{Convergence Rate of Approximation Algorithms}
%     \label{tab:SBM_convergence_rate}
% \end{table*}

Combining Table~\ref{tab:Thoery_bound_convergence_rate}'s and Table~\ref{tab:SBM_communication_cost}'s results, we observe i.i.d. data reduces the gradient variance but increases the communication cost, while the non-i.i.d. setting does the opposite. \emph{Approximation methods via one-hop communication then might be able to balance the convergence rate and communication.} We experimentally validate this intuition in Section~\ref{sec:experiment}'s results, as well as the next appendix section.

\section{Additional Experimental Results}

\subsection{Validation of Theoretical Analysis on Cora Dataset}\label{subsec:validation}
\begin{figure*}[ht]
    \centering
    \includegraphics[width = 0.32\textwidth]{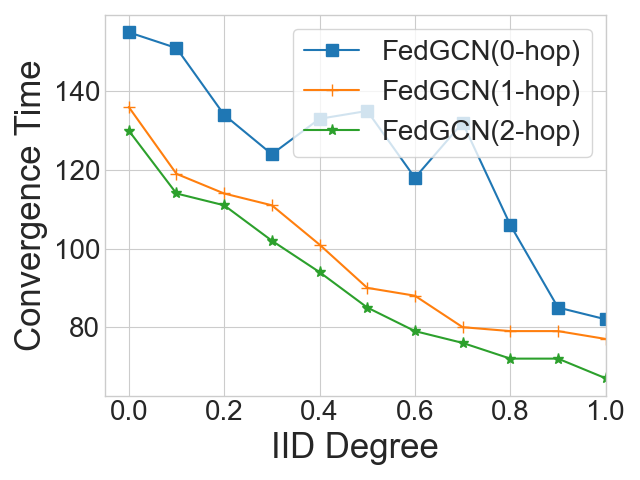}
    \includegraphics[width = 0.315\textwidth]{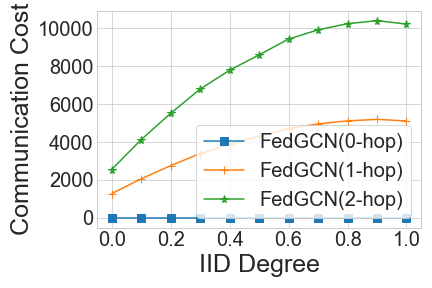}
    \includegraphics[width = 0.32\textwidth]{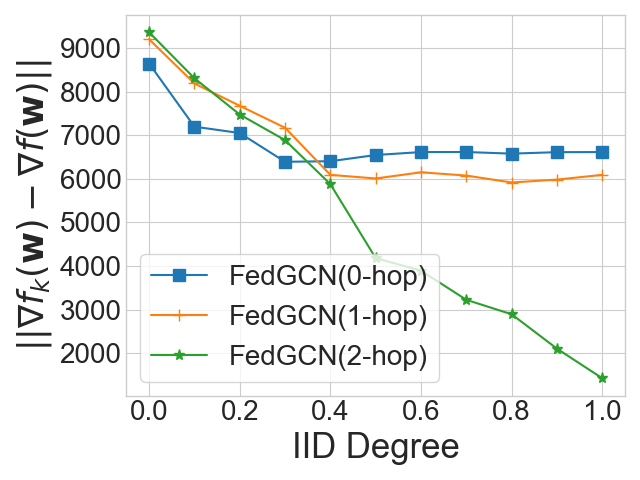}

    \caption{Convergence time (left), communication cost (middle) on Cora, 
    and theoretical convergence upper bound (right, Table~\ref{tab:Thoery_bound_convergence_rate}). FedGCN (1-hop) balances convergence and communication. 
    %The theoretical upper bounds qualitatively hold, with similar trends with the degree of i.i.d. data for 0-hop and 2-hop FedGCN.
    }
    \label{fig:convergence_time_changeIID}
\end{figure*}

We validate the qualitative results in the main theory and~\ref{prop:comm} on the Cora dataset. As shown in Figure \ref{fig:convergence_time_changeIID}, 0-hop FedGCN does not need to communicate but requires a high convergence time. One- and 2-hop FedGCN have similar convergence times, but 1-hop FedGCN needs much less communication. 
The right graph in Figure~\ref{fig:convergence_time_changeIID} shows Table~\ref{tab:Thoery_bound_convergence_rate}'s gradient norm bound for the Cora dataset. We expect these to qualitatively follow the same trends as we increase the fraction of i.i.d. data, since from Theorem~\ref{thm:convergence} the convergence time increases with $\left\|\nabla f_k(\bm{w}_k) - \nabla f(\bm{w})\right\|$. FedGCN (2-hop) and FedGCN (0-hop), as we would intuitively expect, respectively decrease and increase: as the data becomes more i.i.d., FedGCN (0-hop) has more information loss, while FedGCN (2-hop) gains more useful information from cross-client edges. Federated learning also converges faster for i.i.d. data, and we observe that FedGCN (0-hop)'s increase in convergence time levels off for $>80\%$ i.i.d. data.

% \subsection{Convergence Time and Communication Cost on SBM Model.}
% As shown in Figure~\ref{fig:convergence_communicaton_time_changeIID}, FedGCN with 2-hop approximation converges faster but requires more communication cost, which is consistent with the experiments in real datasets.
% \begin{figure}[ht]
%     \centering
%     \includegraphics[width = 0.45\textwidth]{experiments/simulate_convergence_time_changeIID.png}
%     \includegraphics[width = 0.45\textwidth]{experiments/simulate_communication_cost_changeIID.png}

%     \caption{Convergence time and communication cost of methods on data distribution with Stochastic Block Model.}
%     \label{fig:convergence_communicaton_time_changeIID}
% \end{figure}

\subsection{Homomorphic Encryption Microbenchmarking}
\label{FHE_eval}
\begin{table}[ht]
    \footnotesize
    \setlength\tabcolsep{2pt}
\begin{center}
 \begin{tabular}{||c | c||}
 \hline
 \textbf{Scheme} & \textbf{Cheon-Kim-Kim-Song (CKKS)} \\ [0.2ex] 
 \hline\hline
 \textbf{ring dimension} & 4096\\ 
 \hline
  \textbf{security level}  & HEStd\_128\_classic \\ 
 \hline
  \textbf{multi depth}  & 1\\ 
 \hline
  \textbf{scale factor bits}  & 30\\ 
 \hline
 \end{tabular}
\end{center}
 \caption{HE Scheme Parameter Configuration On PALISADE. Multi depth is configured to be 1 for optimal (minimum) maximum possible multiplicative depth in our evaluation. }
 \label{fhe_parameter} 
\end{table}

We implement our HE module using the HE library PALISADE (v1.10.5)~\citet{palisade} with the cryptocontext parameters configuration as in Table~\ref{fhe_parameter}. In our paper, we evaluate the real-number HE scheme, i.e., the Cheon-Kim-Kim-Song (CKKS) scheme~\citet{cheon2017homomorphic}. 
%\yuhang{BGV is able to encrypt booleans and integers without error. CKKS can encrypt real number with little error, which can be ignored in FedAvg. [Cite a paper?]}

\begin{table*}[ht]
\begin{center}
\begin{tabular}{||c | c| c | c | c ||} 
 \hline
Array Size& Plaintext (Bool)& Plaintext (Long \& Double)&	CKKS	& CKKS (Boolean Packing)\\
  \hline\hline
          1k & 1 kB & 8 kB & 266 kB & 266 kB \\ \hline
        10k & 10 kB & 80 kB & 798 kB & 266 kB \\ \hline
        100k & 100 kB & 800 kB & 7 MB & 1 MB \\ \hline
        1M & 1 MB & 8 MB & 70 MB & 8 MB \\ \hline
        100M & 100 MB & 800 MB & 7 GB & 793 MB \\ \hline
        1B & 1 GB & 8 GB & 70 GB & 8 GB \\ \hline
% Key Generation & \\ 
%  \hline\hline

\end{tabular}
\end{center}
 \caption{Communication Cost Comparison between Plaintext and Encryption: Plaintext files are numpy arrays with pickle and ciphertext files are generated under CKKS.}
\label{comm_fhe}
\end{table*}

In our framework, neighboring features (long integers, int64) are securely aggregated under the BGV scheme and local model parameters (double-precision floating-point, float64) are securely aggregated under the CKKS scheme. The microbenchmark results of additional communication overhead can be found in Table~\ref{comm_fhe}. In general, secure computation using HE yields a nearly 15-fold increase in communicational cost compared to insecure communication in a complete view of plaintexts. However, with our Boolean Packing technique, the communication overhead only \textbf{doubles} for a large-size array. 

\section{Assumptions of Proof}
\subsection{Standard Assumptions}
Lipschitz Continuous Gradient, Bounded Global Variability, and Bounded Gradient are standard assumptions for FL analysis~\cite{li2019convergence, yu2019parallel, yang2021achieving, ramezani2021learn,koloskova2022sharper}. For example, \cite{ramezani2021learn} incorporates them as assumption 1, assumption 2, and Eqn. 14 of Appendix B.2 in their paper, 
$$\mathbb{E}[\mathcal{L}(\bar{\theta}^{t+1})]\leq\mathbb{E}[\mathcal{L}(\bar{\theta}^t)]+\mathbb{E}[\langle\nabla\mathcal{L}(\bar{\theta}^t),\bar{\theta}^{t+1}-\bar{\theta}^t\rangle]+\frac{L}{2}\mathbb{E}[\|\bar{\theta}^{t+1}-\bar{\theta}^t\|^2].$$

They are “very standard non-convex, non-iid FL assumptions”, which has been the state-of-the-art FL convergence analysis.

However, we also agree that these non-convex assumptions are still relatively strict, since how to provide FL analysis without such assumptions is still an open problem. 

Bounded Global Variability and Bounded Gradient allow the convergence analysis to accommodate data distribution heterogeneity, a core challenge of FL. 

The bounded gradient assumption in particular holds for certain activation functions, e.g., sigmoid functions, and bounded input features. 
Lipschitz Continuous Gradient is a technical condition on the shape of the loss function that is standard for non-convex analysis. 
It in fact relaxes the assumption of (strongly) convex loss functions that were previously common in analyzing FL and stochastic gradient descent. Our theory is also based on [2]’s convergence result for FedAvg. We hope our theory can open the area of theoretical analysis of federated graph learning.

\subsection{Are these assumptions still valid for the graph?}

For Assumption 5.3 ($\lambda$-Lipschitz Continuous Gradient),

$$\|\nabla f_k(w)-\nabla f_k(v)\|\leq\lambda\|w-v\|, \forall w,v \in\mathbb{R}^d,$$

$w$ and $v$ in the statement of this assumption represent two arbitrary sets of parameter values of the model. In a graph neural network, $w$ represents the concatenation of parameters of each layer, i.e., $[W_1,W_2,...,W_L]$ where $W_l$ is the vectorized weight matrix of the $l$-th layer. For example, $w$ can be the model parameters at the first training iteration, and $v$ can be the model parameters after subsequent training iterations.

The assumption is general for arbitrary $w$ and $v$. In other words, it means that changing the model parameters from $w$ to $v$ will not change the value of the loss function $\nabla f_k$ by more than a constant factor, multiplied by the norm of $\left\|w-v\right\|$. The correlation between $w$ and $v$ will not affect the bound. Intuitively, one might in fact expect a correlation between $w$ and $v$ to make the Lipschitz property more likely to hold, since the loss value is less likely to change much if the new parameter values $(v)$ are correlated with the old parameter values $(w)$.

Assumption 5.4, $\|\nabla f_k(w_t)-\nabla f(w_t)\|\leq\sigma_G$, follows from Assumption 5.5. If the local gradient is bounded, then since the global gradient is the sum of local gradients, it is also bounded. Thus, the difference between local and global gradients will also be bounded.

For Assumption 5.5, 
$\|\nabla f_k(w_t)\|\leq G,$
we agree that the bounded gradient assumption may not always hold. However, it can be shown that this assumption holds for certain activation functions, e.g., sigmoid functions, and bounded input features. 

\subsection{Why do we adopt such assumptions?}

Our analysis is based on~\citep{yu2019parallel}. To the best of our knowledge, all state-of-the-art FL papers, such as~\cite{li2019convergence, yu2019parallel, yang2021achieving, ramezani2021learn,koloskova2022sharper}, make very similar assumptions. Since the main purpose of our paper is not to advance the convergence analysis of FL in general, but rather to show how this analysis applies to federated graph training, we follow these papers' assumptions. We believe that if better FL theory papers emerge that remove one or all assumptions, we can extend our work to this more advanced theory by analyzing the difference between the local gradient and global gradient in the graph setting.

We further believe that, while our exact quantitative convergence bounds may not hold in practice given that some of the theoretical assumptions may be violated, the qualitative insights derived from those bounds may still be valuable. In Figure~\ref{fig:cross_device_cross_silo}, for example, we empirically validate our qualitative observations on how FedGCN's convergence varies with the number of clients and number of hops. We will further emphasize in the paper that the convergence analysis suggests qualitative insights about FedGCN's performance, even if the exact mathematical expressions do not always hold.

\section{Convergence Proof}\label{appen:Convegence}
We first give an example of a 1-layer GCN, then we mainly analyze a 2-layer GCN, which is the most common architecture for graph neural networks. \emph{The intuition of the theory is bounding the difference between the local gradient and global gradient in non-i.i.d settings.} Our analysis also fits any layers of GCN and GraphSage. 

\subsection{Convergence Analysis of 1-layer GCNs}
We first get the gradient of 1-layer GCNs in centralized, 0-hop, and 1-hop cases. Then we provide bounds to approximate the difference between local (0,1-hop) and global gradients.
\subsubsection{Gradient of Centralized GCN (Global Gradient)}
In centralized training, we consider a graph $\mathcal{G}$ with $N$ nodes and $d$-dim feature for each node. The graph can be also represented as the adjacency matrix $\bm{A}$ and the feature matrix $\bm{X}$. Each nodes belongs to a specific class $c$, where the node label matrix can be represented as $\bm{Y}$. We then consider a 1-layer graph convolutional network with model parameter $\bm{W}$ and softmax activation $\phi$, which has the following form

\begin{equation}
    \bm{Z} = \bm{A}\bm{X}\bm{W}.
\end{equation}
We then pass it to the softmax activation
\begin{equation}
    \bm{Q} = \phi (\bm{Z}),
\end{equation}
where 
\begin{equation}
    \bm{Q}_{ic} =  \frac{e^{\bm{Z}_{ic}}}{\sum_{c=1}^C e^{\bm{Z}_{ic}}}.
\end{equation}
$\bm{Q}_{ic}$ is then the model prediction result for node $i$ with specific class $c$.

Let $f(\bm{A}, \bm{X}, \bm{W}, \bm{Y})$ represent the output of the cross-entropy loss, we have
\begin{equation}
    f(\bm{A}, \bm{X}, \bm{W}, \bm{Y}) = -\frac{1}{N} \sum_{i=1}^N \sum_{c=1}^C \bm{Y}_{ic} \log \bm{Q}_{ic}.
\end{equation}

\textbf{Equation 1} Gradient to the input of softmax layer $\frac{\partial f}{\partial \bm{Z}} = \frac{1}{N} (\bm{Q} - \bm{Y})$
\begin{proof}
At first, we calculate the gradient of $f$ given the element $\bm{Z}_{ic}$ of the matrix $\bm{Z}$, $\frac{\partial f}{\partial \bm{Z}_{ic}}$, 

\begin{equation}
\begin{aligned}
\frac{\partial f}{\partial \bm{Z}_{ic}} 
&= \frac{\partial (-\frac{1}{N} \sum_{i=1}^N \sum_{c=1}^C \bm{Y}_{ic} \log \bm{Q}_{ic})}{\partial \bm{Z}_{ic}}\\
&= \frac{\partial (-\frac{1}{N} \sum_{i=1}^N \sum_{c=1}^C \bm{Y}_{ic} \log \frac{e^{\bm{Z}_{ic}}}{\sum_{d=1}^C e^{\bm{Z}_{id}}})}{\partial \bm{Z}_{ic}}\\
&=\frac{\partial (-\frac{1}{N} \sum_{c=1}^C \bm{Y}_{ic} \log \frac{e^{\bm{Z}_{ic}}}{\sum_{d=1}^C e^{\bm{Z}_{id}}})}{\partial \bm{Z}_{ic}}\\
&=-\frac{1}{N}\frac{\partial ( \sum_{c=1}^C \bm{Y}_{ic} \log \frac{e^{\bm{Z}_{ic}}}{\sum_{d=1}^C e^{\bm{Z}_{id}}})}{\partial \bm{Z}_{ic}}\\
&=-\frac{1}{N}  \frac{\partial ( \sum_{c=1}^C (\bm{Y}_{ic} {\bm{Z}_{ic}} - \bm{Y}_{ic} \log {\sum_{d=1}^C e^{\bm{Z}_{id}}}))}{\partial \bm{Z}_{ic}}\\
&=-\frac{1}{N}  (\bm{Y}_{ic} - \frac{\partial ( \sum_{c=1}^C (\bm{Y}_{ic} \log {\sum_{d=1}^C e^{\bm{Z}_{id}}}))}{\partial \bm{Z}_{ic}})\\
&= -\frac{1}{N}  (\bm{Y}_{ic} - \frac{\partial ( \log {\sum_{d=1}^C e^{\bm{Z}_{id}}}))}{\partial \bm{Z}_{ic}})\\
&= -\frac{1}{N}  (\bm{Y}_{ic} - \frac{e^{\bm{Z}_{ic}}}{\sum_{d=1}^C e^{\bm{Z}_{id}}})\\
&= \frac{1}{N}  ( \frac{e^{\bm{Z}_{ic}}}{\sum_{d=1}^C e^{\bm{Z}_{id}}}- \bm{Y}_{ic})\\
&= \frac{1}{N}  (\bm{Q}_{ic}- \bm{Y}_{ic})
\end{aligned}
\end{equation}

Given the property of the matrix, we have

$$\frac{\partial f}{\partial \bm{Z}} = \frac{1}{N} (\bm{Q} - \bm{Y}).$$
\end{proof}

\textbf{Lemma 1} If $\bm{Z} = \bm{A}\bm{X}\bm{B}$, 
$$\frac{\partial f}{\partial \bm{X} } =\bm{A}^T  \frac{\partial f}{\partial \bm{Z} } \bm{B}^T .$$

\textbf{Equation 2} The gradient over the weights of GCN
\begin{equation}
\begin{aligned}
    \frac{\partial {f}}{\partial \bm{W} } =\frac{1}{N} {\bm{X}}^T {\bm{A}}^T (\phi({\bm{A}}{\bm{X}}\bm{W}) - \bm{Y}).\\
\end{aligned}
\end{equation}

\begin{proof}
\begin{equation}
\begin{aligned}
    \frac{\partial {f}}{\partial \bm{W} }
&=({\bm{A}}{\bm{X}})^T \frac{\partial f}{\partial \bm{Z}} \\
&={\bm{X}}^T {\bm{A}}^T \frac{\partial f}{\partial \bm{Z}} \\
&=\frac{1}{N} {\bm{X}}^T {\bm{A}}^T (\bm{Q} - \bm{Y})\\
&=\frac{1}{N} {\bm{X}}^T {\bm{A}}^T (\phi({\bm{A}}{\bm{X}}\bm{W}) - \bm{Y})\\
\end{aligned}
\end{equation}
\end{proof}

\subsubsection{Gradients of local models with 0-hop communication}

We then consider the federated setting. Let $\bm{A}^{N \times N}$ denote the adjacency matrix of all nodes and $\bm{A}_k^{N_k \times N_k}$ denotes the adjacency matrix of the nodes in client $k$. Let $f_k$ represent the local loss function (without communication) of client $k$. Then the local gradient given model parameter $\bm{W}$ is

\begin{equation}
\begin{aligned}
    \frac{\partial {f_k}}{\partial \bm{W} }=\frac{1}{N_k} {\bm{X}}_k^T {\bm{A}}_k^T (\phi({\bm{A}}_k{\bm{X}}_k\bm{W}) - \bm{Y}_k)\\
\end{aligned}
\end{equation}

\subsubsection{Gradients of local models with 1-hop communication}
With 1-hop communication, let $\dot{\bm{A}}_k^{N_k \times |\mathcal{N}_k|}$ denotes the adjacency matrix of the nodes in client $k$ and their 1-hop neighbors ($\mathcal{N}_k$ also includes the current nodes). The output of GCN with 1-hop communication (recovering 1-hop neighbor information) is
\begin{equation}
    \phi (\dot{\bm{A}}_k\dot{\bm{X}}_k\bm{W}).
\end{equation}

The local gradient with 1-hop communication given model parameter $\bm{W}$ is then 
\begin{equation}
\begin{aligned}
    \frac{\partial {\dot{f}_k}}{\partial \bm{W} }=\frac{1}{N_k} \dot{\bm{X}}_k^T \dot{\bm{A}}_k^T (\phi(\dot{\bm{A}}_k\dot{\bm{X}}_k\bm{W}) - \bm{Y}_k)\\
\end{aligned}
\end{equation}

\subsubsection{Bound the difference of local gradient and global gradient}
Assuming each client has an equal number of nodes, we have $N_k = \frac{N}{K}$. The local gradient of 0-hop communication is then
\begin{equation}
\begin{aligned}
    \frac{\partial {f_k}}{\partial \bm{W} }=\frac{K}{N} {\bm{X}}_k^T {\bm{A}}_k^T (\phi({\bm{A}}_k{\bm{X}}_k\bm{W}) - \bm{Y}_k)\\
\end{aligned}
\end{equation}
The difference between the local gradient (0-hop) and the global gradient is then
\begin{equation}
\begin{aligned}
    \| &\frac{\partial {f_k}}{\partial \bm{W} } - \frac{\partial {f}}{\partial \bm{W} }\| \\ &=\| \frac{K}{N} {\bm{X}}_k^T {\bm{A}}_k^T (\phi({\bm{A}}_k{\bm{X}}_k\bm{W}) - \bm{Y}_k) - \frac{1}{N} {\bm{X}}^T {\bm{A}}^T (\phi({\bm{A}}{\bm{X}}\bm{W}) - \bm{Y})\|\\
    &=\frac{1}{N} \| K {\bm{X}}_k^T {\bm{A}}_k^T (\phi({\bm{A}}_k{\bm{X}}_k\bm{W}) - \bm{Y}_k) - {\bm{X}}^T {\bm{A}}^T (\phi({\bm{A}}{\bm{X}}\bm{W}) - \bm{Y})\|\\
    &=\frac{1}{N} \| K {\bm{X}}_k^T {\bm{A}}_k^T \phi({\bm{A}}_k{\bm{X}}_k\bm{W}) - K{\bm{X}}_k^T {\bm{A}}_k^T \bm{Y}_k - {\bm{X}}^T {\bm{A}}^T \phi({\bm{A}}{\bm{X}}\bm{W})  + {\bm{X}}^T {\bm{A}}^T \bm{Y}\|\\
    &\leq \frac{1}{N} (\| K {\bm{X}}_k^T {\bm{A}}_k^T \phi({\bm{A}}_k{\bm{X}}_k\bm{W})  - {\bm{X}}^T {\bm{A}}^T \phi({\bm{A}}{\bm{X}}\bm{W})\|  + \| {\bm{X}}^T {\bm{A}}^T \bm{Y} - K{\bm{X}}_k^T {\bm{A}}_k^T \bm{Y}_k\|)\\
    &= \frac{1}{N} (\| K {\bm{X}}_k^T {\bm{A}}_k^T \phi({\bm{A}}_k{\bm{X}}_k\bm{W})  - {\bm{X}}^T {\bm{A}}^T \phi({\bm{A}}{\bm{X}}\bm{W})\|  + \|K{\bm{X}}_k^T {\bm{A}}_k^T \bm{Y}_k- {\bm{X}}^T {\bm{A}}^T \bm{Y}\|)\\
    &\lesssim \| K {\bm{X}}_k^T {\bm{A}}_k^T \phi({\bm{A}}_k{\bm{X}}_k\bm{W})  - {\bm{X}}^T {\bm{A}}^T \phi({\bm{A}}{\bm{X}}\bm{W})\|  + \|K{\bm{X}}_k^T {\bm{A}}_k^T \bm{Y}_k- {\bm{X}}^T {\bm{A}}^T \bm{Y}\|\\
    \label{eqn:localglobal_real_1layer}
\end{aligned}
\end{equation}

Since model training is to make the model output $\phi({\bm{A}}{\bm{X}}\bm{W})$ close to label matrix $\bm{Y}$, we  provide an upper bound
\begin{equation}
    \|K{\bm{X}}_k^T {\bm{A}}_k^T \bm{Y}_k- {\bm{X}}^T {\bm{A}}^T \bm{Y}\| \lesssim \| K {\bm{X}}_k^T {\bm{A}}_k^T \phi({\bm{A}}_k{\bm{X}}_k\bm{W}) - {\bm{X}}^T {\bm{A}}^T \phi({\bm{A}}{\bm{X}}\bm{W})\|
    \label{eqn:label_model_output_bound_1layer}
\end{equation}
Based on Equation~\ref{eqn:localglobal_real_1layer} and Equation~\ref{eqn:label_model_output_bound_1layer} , we then have

\begin{equation}
\begin{aligned}
    \| \frac{\partial {f_k}}{\partial \bm{W} } - \frac{\partial {f}}{\partial \bm{W} }\| 
    \lesssim \| K {\bm{X}}_k^T {\bm{A}}_k^T \phi({\bm{A}}_k{\bm{X}}_k\bm{W})  - {\bm{X}}^T {\bm{A}}^T \phi({\bm{A}}{\bm{X}}\bm{W}) \|\\
\end{aligned}
\end{equation}

By assuming the function ${\bm{X}}^T {\bm{A}}^T \phi({\bm{A}}{\bm{X}}\bm{W})$ is $\lambda$-smooth w.r.t $ {\bm{X}}^T {\bm{A}}^T {\bm{A}}{\bm{X}}$, we have 
\begin{equation}
\begin{aligned}
    \| \frac{\partial {f_k}}{\partial \bm{W} } - \frac{\partial {f}}{\partial \bm{W} }\| 
    &\lesssim \| K {\bm{X}}_k^T {\bm{A}}_k^T {\bm{A}}_k{\bm{X}}_k\bm{W}  - {\bm{X}}^T {\bm{A}}^T {\bm{A}}{\bm{X}}\bm{W} \|\\
    &\lesssim \| K {\bm{X}}_k^T {\bm{A}}_k^T {\bm{A}}_k{\bm{X}}_k  - {\bm{X}}^T {\bm{A}}^T {\bm{A}}{\bm{X}} \|\\
\end{aligned}
\end{equation}

We can then provide the following bound to compare the local gradient with 0-hop communication and the global gradient given the same model parameter $\bm{w}$
\begin{equation}
\begin{aligned}
    \| \frac{\partial {f_k}}{\partial \bm{w} } - \frac{\partial {f}}{\partial \bm{w} }\| 
    \lesssim \| K {\bm{X}}_k^T {\bm{A}}_k^T {\bm{A}}_k{\bm{X}}_k  - {\bm{X}}^T {\bm{A}}^T {\bm{A}}{\bm{X}} \|,
\end{aligned}
\end{equation}
where $\bm{w}$ is the vectorization of model parameters $\bm{W}$.

Similarly, let $\dot{f}_k$ represent the loss function with 1-hop communication, the difference between the local gradient with 1-hop communication and the global gradient is

\begin{equation}
\begin{aligned}
\|\frac{\partial \dot{f}_k}{\partial \bm{w}} - \frac{\partial f}{\partial \bm{w}}\| \lesssim \lambda \| K \dot{\bm{X}}_k^T \dot{\bm{A}}_k^T  \dot{\bm{A}}_k\dot{\bm{X}}_k - \bm{X}^T \bm{A}^T \bm{A}\bm{X}\|.
\end{aligned}
\end{equation}

\subsection{Convergence Analysis of 2-layer GCNs}
Based on the same idea in 1-layer GCNs, we provide the convergence analysis of 2-layer GCNs. We first derive the gradient of 2-layer GCNs in centralized, 0-hop, 1-hop, and 2-hop cases. Then we provide bounds to approximate the difference between local (0, 1, 2-hop) and global gradients. Based on the Stochastic Block Model, we are able to quantify the difference.

\subsubsection{Gradient of Centralized GCN (Global Gradient)}
Based on the analysis of 1-layer GCNs, for graph $\mathcal{G}$ with adjacency matrix $\bm{A}$ and feature matrix $\bm{X}$ in clients, we consider a 2-layer graph convolutional network with ReLU activation for the first layer, Softmax activation for the second layer, and cross-entropy loss, which has the following form
\begin{equation}
    \bm{Z} = \bm{A}\phi_1(\bm{A}\bm{X}\bm{W}_1)\bm{W}_2,
\end{equation}
\begin{equation}
    \bm{Q} = \phi_2 (\bm{Z}),
\end{equation}
where 
\begin{equation}
    \bm{Q}_{ic} =  \frac{e^{\bm{Z}_{ic}}}{\sum_{d=1}^C e^{\bm{Z}_{id}}}
\end{equation}
The objective function is
\begin{equation}
    f(\bm{A}, \bm{X}, \bm{W}_1, \bm{W}_2, \bm{Y}) = -\frac{1}{N} \sum_{i=1}^N \sum_{c=1}^C \bm{Y}_{ic} \log \bm{Q}_{ic}.
\end{equation}

We then show how to calculate the gradient $\nabla f (\mathbf{w})= [\frac{\partial f}{\partial \bm{W}_1}, \frac{\partial f}{\partial \bm{W}_2}] $.

\textbf{Equation 1} $\frac{\partial f}{\partial \bm{Z}} = \frac{1}{N} (\bm{Q} - \bm{Y})$

\textbf{Equation 2} The gradient over the weights of the second layer
\begin{equation}
\begin{aligned}
    \frac{\partial f}{\partial \bm{W}_2 }
&=\frac{1}{N} (\phi_1(\bm{W}_1^T {\bm{X}}^T {\bm{A}}^T )) {\bm{A}}^T (\phi_2 ({\bm{A}}\phi_1({\bm{A}}{\bm{X}}\bm{W}_1)\bm{W}_2) - \bm{Y})
\end{aligned}
\end{equation}

\begin{proof}
\begin{equation}
\begin{aligned}
    \frac{\partial {f}}{\partial \bm{W}_2 }
&=({\bm{A}}\phi_1({\bm{A}}{\bm{X}}\bm{W}_1))^T \frac{\partial f}{\partial \bm{Z}} \\
&=(\phi_1({\bm{A}}{\bm{X}}\bm{W}_1))^T {\bm{A}}^T \frac{\partial f}{\partial \bm{Z}} \\
&=\frac{1}{N} (\phi_1({\bm{A}}{\bm{X}}\bm{W}_1))^T {\bm{A}}^T (\bm{Q} - \bm{Y})\\
&=\frac{1}{N} (\phi_1({\bm{A}}{\bm{X}}\bm{W}_1))^T {\bm{A}}^T (\phi_2 ({\bm{A}}\phi_1({\bm{A}}{\bm{X}}\bm{W}_1)\bm{W}_2) - \bm{Y})\\
&=\frac{1}{N} (\phi_1(\bm{W}_1^T {\bm{X}}^T {\bm{A}}^T )) {\bm{A}}^T (\phi_2 ({\bm{A}}\phi_1({\bm{A}}{\bm{X}}\bm{W}_1)\bm{W}_2) - \bm{Y})\\
\end{aligned}
\end{equation}
\end{proof}

\textbf{Equation 3} The gradient over the weights of the first layer.
\begin{equation}
    \frac{\partial f}{\partial \bm{W}_1} = \frac{1}{N}  ({\bm{A}}\phi'_1({\bm{A}}{\bm{X}}\bm{W}_1) {\bm{A}}{\bm{X}})^T  (\phi_2 ({\bm{A}}\phi_1({\bm{A}}{\bm{X}}\bm{W}_1)\bm{W}_2)-\bm{Y}) \bm{W}_2^T
\end{equation}
\begin{proof}
\begin{equation}
\begin{aligned}
\frac{\partial {f}}{\partial \bm{W}_1 }&=  ({\bm{A}}\phi'_1({\bm{A}}{\bm{X}}\bm{W}_1) {\bm{A}}{\bm{X}})^T \frac{\partial f}{\partial \bm{Z}} \bm{W}_2^T\\
     &=   ({\bm{A}}\phi'_1({\bm{A}}{\bm{X}}\bm{W}_1) {\bm{A}}{\bm{X}})^T  \frac{\partial f}{\partial \bm{Z}} \bm{W}_2^T\\
     &=  \frac{1}{N}  ({\bm{A}}\phi'_1({\bm{A}}{\bm{X}}\bm{W}_1) {\bm{A}}{\bm{X}})^T  (\bm{Q}-\bm{Y}) \bm{W}_2^T\\
     &=  \frac{1}{N}  ({\bm{A}}\phi'_1({\bm{A}}{\bm{X}}\bm{W}_1) {\bm{A}}{\bm{X}})^T  (\phi_2 ({\bm{A}}\phi_1({\bm{A}}{\bm{X}}\bm{W}_1)\bm{W}_2)-\bm{Y}) \bm{W}_2^T
\end{aligned}
\end{equation}
\end{proof}
\subsubsection{Gradient of local models (0, 1, 2-hop)}

For client $k$ with local adjacency matrix $\bm{A}_k$, let $\dot{\bm{A}}_k^{n \times |\mathcal{N}_k|}$ denotes the adjacency matrix of the current nodes with complete edge information form their 1-hop neighbors ($\mathcal{N}_k$ also includes the current nodes), and $\ddot{\bm{A}}_k^{|\mathcal{N}_k| \times |\mathcal{N}^2_k|}$ denotes the adjacency matrix of nodes with complete edge information form their 2-hop neighbors ($\mathcal{N}^2_k$ also includes the current nodes and 1-hop neighbors).

The output of GCN without communication is
\begin{equation}
    \phi_2( {\bm{A}}_k\phi_1({\bm{A}}_k{\bm{X}}_k\bm{W}_1)\bm{W}_2).
\end{equation}

The output of GCN with 1-hop communication is
\begin{equation}
    \phi_2( {\bm{A}}_k\phi_1(\dot{\bm{A}}_k\dot{\bm{X}}_k\bm{W}_1)\bm{W}_2).
\end{equation}
The output of GCN with 2-hop communication is
\begin{equation}
    \phi_2( \dot{\bm{A}}_k\phi_1(\ddot{\bm{A}}_k\ddot{\bm{X}}_k\bm{W}_1)\bm{W}_2).
\end{equation}

For 2-layer GCNs, output with 2-hop communication is the same as the centralized model.

The gradient of GCNs with 2-hop communication (recover the 2-hop neighbor information) over the weights of the first layer is then
\begin{equation}
    \frac{\partial \ddot{f}_k}{\partial \bm{W}_1} = \frac{1}{N_k}  (\dot{\bm{A}}_k\phi'_1(\ddot{\bm{A}}_k\ddot{\bm{X}}_k\bm{W}_1) \ddot{\bm{A}}_k\ddot{\bm{X}}_k)^T  (\phi_2 (\dot{\bm{A}}_k\phi_1(\ddot{\bm{A}}_k\ddot{\bm{X}}_k\bm{W}_1)\bm{W}_2)-\bm{Y}_k) \bm{W}_2^T.
\end{equation}

\subsubsection{Bound the difference of local gradient and global gradient}

Assuming each client has an equal number of nodes, we have $N_k = \frac{N}{K}$. Based on the same process in 1-layer case, we can then provide the following approximations between the local model and the global model.

The difference between the local gradient without communication and the global gradient is
\begin{equation}
\begin{aligned}
\|\frac{\partial {f}_k}{\partial \bm{w}} - \frac{\partial f}{\partial \bm{w}}\| \lesssim \|K {\bm{X}}^T_k {\bm{A}}^T_k  {\bm{A}}^T_k  {\bm{A}}_k {\bm{A}}_k {\bm{X}}_k - \bm{X}^T \bm{A}^T  \bm{A}^T  \bm{A}\bm{A}\bm{X}\|.
\end{aligned}
\end{equation}

The difference between the local gradient with 1-hop communication and the global gradient is
\begin{equation}
\begin{aligned}
\|\frac{\partial \dot{f}_k}{\partial \bm{w}} - \frac{\partial f}{\partial \bm{w}}\| \lesssim \|K \dot{\bm{X}}_k^T \dot{\bm{A}}_k^T  {\bm{A}}_k^T  {\bm{A}}_k\dot{\bm{A}}_k\dot{\bm{X}}_k - \bm{X}^T \bm{A}^T  \bm{A}^T  \bm{A}\bm{A}\bm{X}\|.
\end{aligned}
\end{equation}

The difference between the local gradient with 2-hop communication and the global gradient is
\begin{equation}
\begin{aligned}
\|\frac{\partial \ddot{f}_k}{\partial \bm{w}} - \frac{\partial f}{\partial \bm{w}}\| \lesssim \|K \ddot{\bm{X}}_k^T \ddot{\bm{A}}_k^T  \dot{\bm{A}}_k^T  \dot{\bm{A}}_k\ddot{\bm{A}}_k\ddot{\bm{X}}_k - \bm{X}^T \bm{A}^T  \bm{A}^T  \bm{A}\bm{A}\bm{X}\|.
\end{aligned}
\end{equation}

With more communication, the local gradient gets closer to the global gradient.

\subsection{Analysis on Stochastic Block Model with Node Features}
To better quantify the difference, we can analyze it on generated graphs, the Stochastic Block Model.
\subsubsection{Preliminaries}

Assume the node feature vector $\bm{x}$ follows the Gaussian distribution with linear projection $\bm{H}$ of node label $\bm{y}$,

\begin{equation}
    \bm{x} \sim \mathcal{N} (\bm{H} \bm{y} , \mathbf{\sigma}),
\end{equation}
we then have the expectation of the feature matrix
\begin{equation}
    E(\bm{X}) = E(\bm{Y} \bm{H} ^T).
\end{equation}

According to the Stochastic Block Model, we have
\begin{equation}
\begin{aligned}
E(\bm{A}) = \bm{P}=\bm{Y} \bm{B} \bm{Y}^T.
\end{aligned}
\end{equation}

\subsubsection{Quantify the gradient difference}

Based on above results, the expectation of the global gradient given the label matrix $\bm{Y}$ is
\begin{equation}
\begin{aligned}
E(\bm{X}^T \bm{A}^T  \bm{A}^T  \bm{A}\bm{A}\bm{X} | \bm{Y}) = \bm{H} {\bm{Y}}^T {\bm{Y}}\bm{B} {\bm{Y}}^T  {\bm{Y}}\bm{B} \bm{Y}^T  \bm{Y}\bm{B} {\bm{Y}}^T {\bm{Y}}\bm{B} {\bm{Y}}^T{\bm{Y}} \bm{H}^T.
\end{aligned}
\end{equation}

Notice that $\bm{Y}^T\bm{Y}$ is counting the number of nodes belonging to each class. Based on this observation, we can better analyze the data distribution.

For adjacency matrix without communication
\begin{equation}
\begin{aligned}
E({\bm{A}}_k) = \bm{Y}_k\bm{B} {\bm{Y}}_k^T.
\end{aligned}
\end{equation}

The expectation of the former gradient given the label matrix $\bm{Y}$ is then
\begin{equation}
\begin{aligned}
E(\bm{X}_k^T \bm{A}_k^T  \bm{A}_k^T  \bm{A}_k\bm{A}_k\bm{X}_k | \bm{Y}) = \bm{H} {\bm{Y}}_k^T {\bm{Y}}_k\bm{B} {\bm{Y}}_k^T  {\bm{Y}}_k\bm{B} \bm{Y}_k^T  \bm{Y}_k\bm{B} {\bm{Y}}_k^T {\bm{Y}}_k\bm{B} {\bm{Y}}_k^T{\bm{Y}}_k \bm{H}^T.
\end{aligned}
\end{equation}

For adjacency matrix with 1-hop communication

\begin{equation}
\begin{aligned}
E(\dot{\bm{A}}_k) = \bm{Y}_k\bm{B} \dot{\bm{Y}}_k^T
\end{aligned}
\end{equation}

The expectation of the former gradient with 1-hop communication given the label matrix $\bm{Y}$ is then
\begin{equation}
\begin{aligned}
E(\dot{\bm{X}}_k^T \dot{\bm{A}}_k^T  \bm{A}_k^T  \bm{A}_k\dot{\bm{A}}_k\dot{\bm{X}}_k | & \bm{Y})
\\ &= \bm{H} \dot{\bm{Y}}_k^T \dot{\bm{Y}}_k\bm{B} {\bm{Y}}_k^T  {\bm{Y}}_k\bm{B} \bm{Y}_k^T  \bm{Y}_k\bm{B} {\bm{Y}}_k^T {\bm{Y}}_k\bm{B} \dot{\bm{Y}}_k^T \dot{\bm{Y}}_k \bm{H}^T.
\end{aligned}
\end{equation}

For adjacency matrix with 2-hop communication
\begin{equation}
\begin{aligned}
E(\ddot{\bm{A}}) = \dot{\bm{Y}}\bm{B} \ddot{\bm{Y}}^T.
\end{aligned}
\end{equation}

The expectation of the former gradient with 1-hop communication given the label matrix $\bm{Y}$ is then
\begin{equation}
\begin{aligned}
E(\ddot{\bm{X}}_k^T \ddot{\bm{A}}_k^T  \dot{\bm{A}}_k^T  \dot{\bm{A}}_k \ddot{\bm{A}}_k \ddot{\bm{X}}_k | &\bm{Y}) 
\\ &=  \bm{H} \ddot{\bm{Y}}_k^T \ddot{\bm{Y}}_k\bm{B} \dot{\bm{Y}}_k^T  \dot{\bm{Y}}_k\bm{B} \bm{Y}_k^T  \bm{Y}_k\bm{B} \dot{\bm{Y}}_k^T \dot{\bm{Y}}_k\bm{B} \ddot{\bm{Y}}_k^T\ddot{\bm{Y}}_k \bm{H}.
\end{aligned}
\end{equation}

The difference in gradient can then be written as

\begin{equation}
\begin{aligned}
\|\frac{\partial \ddot{f}_k}{\partial \bm{w}} - \frac{\partial f}{\partial \bm{w}}\| 
\leq \lambda \|K \ddot{\bm{Y}}_k^T \ddot{\bm{Y}}_k\bm{B} \dot{\bm{Y}}_k^T  \dot{\bm{Y}}_k\bm{B} \bm{Y}_k^T  \bm{Y}_k\bm{B} \dot{\bm{Y}}_k^T \dot{\bm{Y}}_k\bm{B} \ddot{\bm{Y}}_k^T\ddot{\bm{Y}}_k \\ - {\bm{Y}}^T {\bm{Y}}\bm{B} {\bm{Y}}^T  {\bm{Y}}\bm{B} \bm{Y}^T  \bm{Y}\bm{B} {\bm{Y}}^T {\bm{Y}}\bm{B} {\bm{Y}}^T{\bm{Y}} \|
\end{aligned}
\end{equation}

Notice that ${\bm{Y}}_k^T {\bm{Y}}_k$ is counting the number of nodes in client $k$ belonging to each class, $\dot{\bm{Y}}_k^T \dot{\bm{Y}}_k$ and $\ddot{\bm{Y}}_k^T \ddot{\bm{Y}}_k$ are respectively counting the number of 1-hop and 2-hop neighbors of nodes in client $k$ belonging to each class. It can be decomposed as 
\begin{equation}
    {\bm{Y}}_k^T {\bm{Y}}_k = N_k \bm{p}_k,
\end{equation}

We then have
\begin{equation}
\begin{aligned}
\|\frac{\partial \ddot{f}_k}{\partial \bm{w}} - \frac{\partial f}{\partial \bm{w}}\| 
&\lesssim \|K  \mathcal{N}_{\mathcal{V}_k}^2 \bm{p}_k \mathcal{N}_{\mathcal{V}_k}^1 \bm{p}_k N_k \bm{p}_k  \mathcal{N}_{\mathcal{V}_k}^1 \bm{p}_k \mathcal{N}_{\mathcal{V}_k}^2 \bm{p}_k - N^5\bm{p}^5\| \|\bm{B}^4\|\\
&\lesssim \|K  N_k (\mathcal{N}_{\mathcal{V}_k}^1)^2 (\mathcal{N}_{\mathcal{V}_k}^2)^2 (\bm{p}_k)^5 - N^5\bm{p}^5\|\\
&\lesssim \|(K  N_k (\mathcal{N}_{\mathcal{V}_k}^1)^2 (\mathcal{N}_{\mathcal{V}_k}^2)^2 - N^5) (\bm{p}_k)^5\| + N^5 \|(\bm{p}_k)^5 - \bm{p}^5\|
\end{aligned}
\end{equation}

$(K  N_k (\mathcal{N}_{\mathcal{V}_k}^1)^2 (\mathcal{N}_{\mathcal{V}_k}^2)^2 - N^5)$ evaluates the difference between the number of nodes with communication in local client and the number of nodes in total. $\|(\bm{p}_k)^5 - \bm{p}^5\|$ evaluates the difference between local distribution and global distribution. The second term can be bounded by 
\begin{equation}
    N^5 \|(\bm{p}_k)^5 - \bm{p}^5\| \leq N^5 {(1 - \frac{1}{C})}^{\frac{5}{2}} (1-p)^5.
\end{equation}

We then work on bounding the first term.

\subsection{Number of 1-hop and 2-hop neighbors for clients}
We need to get the number of 1-hop neighbors $\mathcal{N}_{\mathcal{V}_k}^1$ and 2-hop neighbors $\mathcal{N}_{\mathcal{V}_k}^2$ in both i.i.d and non-i.i.d cases.
\subsubsection{Number of 1-hop and 2-hop neighbors in i.i.d}

Recall the the definition of SBM~\ref{appen:SBM}, the edge between two nodes is independent of other edges. For node $i$ in other clients, the probability that it has at least one edge with the nodes in client $i$
\begin{equation}
\begin{aligned}
1 - (1-\alpha)^{\frac{N}{C K}} (1-\mu \alpha)^{\frac{(C-1)N}{C K}}
\end{aligned}
\end{equation}

The expectation of 1-hop neighbor (including nodes in local client)

\begin{equation}
\begin{aligned}
\frac{N}{K} + \frac{K-1}{K} N (1 - (1-\alpha)^{\frac{N}{CK}} (1-\mu \alpha)^{\frac{(C-1)N}{CK}}) 
& \approx  \frac{N}{K} + \frac{K-1}{K} N (1 - (1-\alpha \frac{N}{C K} ) (1-\mu \alpha \frac{(C-1)N}{CK} )) \\
& \approx  \frac{N}{K} + \frac{K-1}{K} N (1 - (1 - \alpha \frac{N}{CK} - \mu \alpha \frac{(C-1)N}{CK} ))\\
& = \frac{N}{K} +  \frac{K-1}{K} N (\alpha \frac{N}{CK} + \mu \alpha \frac{(C-1)N}{CK} )\\
& = \frac{N}{K} (1 + (K-1) (\alpha \frac{N}{CK} + \mu \alpha \frac{(C-1)N}{CK} ))\\
\end{aligned}
\end{equation}

Notice that it is $(1 + (K-1) (\alpha \frac{N}{CK} + \mu \alpha \frac{(C-1)N}{CK} )$ propotional to the number of local nodes.

Similarly, approximated expectation of 2-hop neighbor (including nodes in local client). This approximation is provided based on that in expectation there is no label distribution shift between 2-hop nodes and 1-hop nodes.

\begin{equation}
\begin{aligned}
\frac{N}{K} (1 + (K-1) (\alpha \frac{N}{CK} + \mu \alpha \frac{(C-1)N}{CK} ))^2
\end{aligned}
\end{equation}

\subsubsection{Number of 1-hop and 2-hop neighbors in non-i.i.d.}
The expectation of 1-hop neighbor (including nodes in local client)
\begin{equation}
\begin{aligned}
\frac{N}{K} +\frac{K-1}{K} N (1 - (1-\mu \alpha)^{\frac{N}{K}}) \\
\approx \frac{N}{K} (1 + \mu\alpha \frac{K-1}{K} N )
\end{aligned}
\end{equation}
Approximated expectation of 2-hop neighbor (Including nodes in local client). 
\begin{equation}
\begin{aligned}
\frac{N}{K} (1 + \mu\alpha \frac{K-1}{K} N )^2\\
\end{aligned}
\end{equation}

\subsubsection{Number of 1-hop and 2-hop neighbors in non-i.i.d.}
The expectation of 1-hop neighbor (including nodes in local client)
\begin{equation}
\begin{aligned}
\frac{N}{K} + \frac{K-1}{K} N (1 - (1-\alpha)^{\frac{Np}{C K}}(1-\mu \alpha)^{\frac{N(C-p)}{C K}}) & \approx \frac{N}{K} +  \frac{K-1}{K} N (\alpha \frac{N}{CK} ( (1 -\mu) p + \mu C))\\
&= \frac{N}{K} +  \frac{K-1}{K} N (\alpha \frac{N}{CK}  (1 -\mu) p + \alpha \frac{N}{CK}  \mu C)\\
&= \frac{N}{K} (1 + (K-1) (\alpha \frac{N}{CK}  (1 -\mu) p + \mu \alpha \frac{N}{K}  ))\\
\end{aligned}
\end{equation}
Approximated expectation of 2-hop neighbor (including nodes in local client)

\begin{equation}
\begin{aligned}
\frac{N}{K} (1 + (K-1) (\alpha \frac{N}{CK}  (1 -\mu) p + \mu \alpha \frac{N}{K}  ))^2\\
\end{aligned}
\end{equation}

\subsection{Data Distribution with Labels}

We assume each label has the same number of nodes. Each client $k$ has the same number of nodes $N_k = \frac{N}{K}$.

For global label distribution, we have
\begin{equation}
\begin{aligned}
 \bm{p} = diag(\frac{1}{C},...,\frac{1}{C})
\end{aligned}
\end{equation}

\subsubsection{i.i.d}
The local label distribution is the same as the global distribution in the i.i.d condition.
\begin{equation}
\begin{aligned}
 \bm{p}_k = diag(\frac{1}{C},...,\frac{1}{C})
\end{aligned}
\end{equation}

\textbf{For local gradient without communication and global gradient,}

\begin{equation}
\begin{aligned}
\|\frac{\partial {f}_k}{\partial \bm{w}} - \frac{\partial f}{\partial \bm{w}}\| 
&\lesssim \| (K (N_k)^5 -N^5 )   diag(\frac{1}{C},...,\frac{1}{C}) ^5\bm{B}^4\| \\
&\lesssim (1 - K \frac{(N_k)^5}{N^5} )N^5 \|    diag(\frac{1}{C},...,\frac{1}{C}) ^5\bm{B}^4\| \\
&\lesssim (1 - K \frac{(N_k)^5}{N^5} )\frac{N^5}{C^5}  \|   \bm{B}^4\| \\
&\lesssim  (1 - \frac{1}{K^4}) \frac{N^5}{C^5}   \|\bm{B}^4\| \\
\end{aligned}
\end{equation}

\textbf{For local gradient with 1-hop communication and global gradient,}

% \begin{equation}
% \begin{aligned}
%  \dot{\bm{Y}}_{k}  \dot{\bm{Y}}_{k}^T = |\mathcal{N}_k| diag(\frac{1}{C},...,\frac{1}{C})
% \end{aligned}
% \end{equation}

\begin{equation}
\begin{aligned}
\|\frac{\partial \dot{f}_k}{\partial \bm{w}} - \frac{\partial f}{\partial \bm{w}}\| &\lesssim \| (K (N_k)^3 |\mathcal{N}_k|^2 -N^5 )   diag(\frac{1}{C},...,\frac{1}{C}) ^5\bm{B}^4\| \\
&\lesssim \| (K \frac{N^3}{K^3} (\frac{N}{C} +  \frac{C-1}{C} N (\alpha \frac{N}{C} + \mu \alpha \frac{(C-1)N}{CK} ))^2-N^5)   diag(\frac{1}{C},...,\frac{1}{C}) ^5\bm{B}^4\| \\
&\lesssim \| (C \frac{N^5}{C^5} (1 +  (C-1) (\alpha \frac{N}{C} + \mu \alpha \frac{(C-1)N}{CK} ))^2-N^5)   diag(\frac{1}{C},...,\frac{1}{C}) ^5\bm{B}^4\| \\
&\lesssim  (1 - \frac{1}{C^4}(1 +  (C-1) (\alpha \frac{N}{C} + \mu \alpha \frac{(C-1)N}{CK} ))^2) \frac{N^5}{C^5}   \|\bm{B}^4\| \\
\end{aligned}
\end{equation}

\textbf{For local gradient with 2-hop communication and global gradient,}

% \begin{equation}
% \begin{aligned}
%  \ddot{\bm{Y}}_{k}  \ddot{\bm{Y}}_{k}^T = |\mathcal{N}_k^2| diag(\frac{1}{C},...,\frac{1}{C})
% \end{aligned}
% \end{equation}

\begin{equation}
\begin{aligned}
\|\frac{\partial \ddot{f}_k}{\partial \bm{w}} - \frac{\partial f}{\partial \bm{w}}\| &\lesssim \| (K (N_k) |\mathcal{N}_k|^2 |\mathcal{N}_k^2|^2 -N^5 )   diag(\frac{1}{C},...,\frac{1}{C}) ^5\bm{B}^4\| \\
&\lesssim \| (K \frac{N}{C} (\frac{N}{C} +  \frac{C-1}{C} N (\alpha \frac{N}{C} + \mu \alpha \frac{(C-1)N}{CK} ))^2  \\ & \quad ((\frac{N}{C} +  \frac{C-1}{C} N (\alpha \frac{N}{C} + \mu \alpha \frac{(C-1)N}{CK} ))^2)^2-N^5)   diag(\frac{1}{C},...,\frac{1}{C}) ^5\bm{B}^4\| \\
&\lesssim \| (K \frac{N^5}{C^5} (1 +  (C-1) (\alpha \frac{N}{C} + \mu \alpha \frac{(C-1)N}{CK} ))^6-N^5)   diag(\frac{1}{C},...,\frac{1}{C}) ^5\bm{B}^4\| \\
&\lesssim  (1 - \frac{1}{C^4}(1 +  (C-1) (\alpha \frac{N}{C} + \mu \alpha \frac{(C-1)N}{CK} ))^6) \frac{N^5}{C^5}   \|\bm{B}^4\| \\
\end{aligned}
\end{equation}

For non-i.i.d, we can simply replace the number of 1-hop and 2-hop neighbors.

\section{Communication Cost under SBM}\label{appen:comm}
Assume the number of clients $K$ is equal to the number of label types in the graph $G$. $d$ represents the dimension of the node feature and $N$ represents the number of nodes. Table~\ref{tab:communication_cost_fedgcn_BDS_GCN} shows the communication cost of FedGCN and BDS-GCN~\cite{wan2022bns}. Distributed training methods like BDS-GCN requires communication per local update, which makes the communication cost increase linearly with the number of global training round $T$ and number of local updates $E$. FedGCN only requires low communication cost at the initial step. %Table~\ref{tab:communication_cost_fedgcn_sbm} shows the communication cost of FedGCN under different data distributions.
\begin{table}[ht]
%\fontsize{8.6pt}{10.32pt}\selectfont
    \centering
    \begin{tabular}{c|c|c|c}
    \hline
        Methods & 1-hop & $L$-hop & BDS-GCN\\
    \hline
        Generic Graph &  $C_1 + N d$ & $C_1 + \sum_{k=1}^{K} |\mathcal{N}_k^{L-1}| d$ & $L T E \rho d \sum_{k=1}^{K} |\mathcal{N}_k^{1} / \mathcal{V}_k|$\\
    \hline
    \end{tabular}
    \caption{Communication costs of FedGCN and BDS-GCN on the generic graph. BDS-GCN requires communication at every local update. }
    \label{tab:communication_cost_fedgcn_BDS_GCN}
\end{table}

\subsection{Server Aggregation}
We consider the communication cost of node $i$ in client $c(i)$. For node $i$, the server needs to receive messages from $c(i)$ (note that $c(i)$ needs to send the local neighbor aggregation) and other clients containing the neighbors of node $i$.

\subsubsection{Non-i.i.d.}
The possibility that there is no connected node in client $j$ for node $i$ is 
\begin{equation}
    (1-\mu \alpha)^{\frac{N}{K}}.
\end{equation}

The possibility that there is at least one connected node in client $j$ for node $i$ is
\begin{equation}
    1 - (1-\mu \alpha)^{\frac{N}{K}}.
\end{equation}

The number of clients that node $i$ needs to communicate with is 
\begin{equation}
   1 + (K-1)(1 - (1-\mu \alpha)^{\frac{N}{K}}).
\end{equation}

The communication cost of $N$ nodes is 
\begin{equation}
    N (1 + (K-1)(1 - (1-\mu \alpha)^{\frac{N}{K}}))d.
\end{equation}

\textbf{1-order Approximation}
To better understand the communication cost, we can expand the form to provide a 1-order approximation
\begin{equation}
(1-\mu \alpha)^{\frac{N}{K}} \approx  1- \mu \alpha {\frac{N}{K}}
\end{equation}

Possibility that there is no connected node in client $j$ for node $i$ is 
\begin{equation}
    1 - (1-\mu \alpha)^{\frac{N}{K}} \approx  1 - 1 +  \mu \alpha {\frac{N}{K}} = \mu \alpha {\frac{N}{K}}.
\end{equation}

The number of clients that node $i$ needs to communicate with is then

\begin{equation}
\begin{aligned}
    1 + (K-1)(1 - (1-\mu \alpha)^{\frac{N}{K}}) \approx 1 + (K-1)\mu \alpha {\frac{N}{K}}.
\end{aligned}
\end{equation}

\subsubsection{i.i.d.}

The possibility that there is no connected node in client $j$ for node $i$ is

\begin{equation}
    (1-\alpha)^{\frac{N}{CK}}(1-\mu \alpha)^{\frac{(C-1)N}{CK}}.
\end{equation}
The possibility that there is at least one connected node in client $j$ for node $i$ is 
\begin{equation}
    1 - (1-\alpha)^{\frac{N}{CK}}(1-\mu \alpha)^{\frac{(C-1)N}{CK}}.
\end{equation}

The number of clients that node $i$ needs to communicate with are 
\begin{equation}
   1 + (C-1)(1 - (1-\alpha)^{\frac{N}{CK}}(1-\mu \alpha)^{\frac{(C-1)N}{CK}}).
\end{equation}

Node $i$ needs to communicate with more clients in i.i.d. than the case in non-i.i.d.

The communication cost of $N$ nodes is 
\begin{equation}
    N (1 + (C-1)(1 - (1-\alpha)^{\frac{N}{CK}}(1-\mu \alpha)^{\frac{(C-1)N}{CK}}))d.
\end{equation}

\textbf{1-order Approximation}

The number of clients that node $i$ needs to communicate with is then

\begin{equation}
\begin{aligned}
    1 - (1-\alpha)^{\frac{N}{CK}}(1-\mu \alpha)^{\frac{(C-1)N}{CK}} 
    &\approx 1- (1- \alpha \frac{N}{CK}) ( 1- \mu \alpha \frac{(C-1)N}{CK}) \\
    &= 1- (1- \alpha \frac{N}{CK} - \mu \alpha \frac{(C-1)N}{CK} + \alpha \frac{N}{CK} \mu \alpha \frac{(C-1)N}{CK})\\
    &= \alpha \frac{N}{CK} + \mu \alpha \frac{(C-1)N}{CK} - \alpha \frac{N}{CK} \mu \alpha \frac{(C-1)N}{CK}\\
    &\approx \alpha \frac{N}{CK} + \mu \alpha \frac{(C-1)N}{CK}.
\end{aligned}
\end{equation}

Therefore, the expression estimates the number of clients that node $i$ needs to communicate with and its own client gives

\begin{equation}
    (1 + (C-1) (\alpha \frac{N}{CK} + \mu \alpha \frac{(C-1)N}{CK})).
\end{equation}

\subsubsection{Non-i.i.d.}

Similarly, let $\bm{p}$ denote the percent of i.i.d., we then have the communication cost
\begin{equation}N(1 + (C-1)(1 - (1-\alpha)^{\frac{Np}{CK}}(1-\mu \alpha)^{\frac{N(C-p)}{CK}}))d.\end{equation}

\textbf{1-order Approximation}

The number of clients that node $i$ needs to communicate with is then

\begin{equation}
\begin{aligned}
    1 - (1-\alpha)^{\frac{Np}{CK}}(1-\mu \alpha)^{\frac{N(C-p)}{CK}}
    &\approx 1- (1- \alpha \frac{Np}{CK}) ( 1- \mu \alpha \frac{N(C-p)}{CK}) \\
    &=  1- (1 - \alpha \frac{Np}{CK} - \mu \alpha \frac{N(C-p)}{CK} + \alpha \frac{Np}{CK} \mu \alpha \frac{N(C-p)}{CK}\\
    &=  \alpha \frac{Np}{CK} + \mu \alpha \frac{N(C-p)}{CK} - \alpha \frac{Np}{CK} \mu \alpha \frac{N(C-p)}{CK}\\
    &\approx \alpha \frac{Np}{CK} + \mu \alpha \frac{N(C-p)}{CK}\\
    &= \alpha \frac{N}{CK} (p + \mu {(C-p))}\\
    &= \alpha \frac{N}{CK} (p -\mu p + \mu C )\\
    &= \alpha \frac{N}{CK} ( (1 -\mu) p + \mu C).
\end{aligned}
\end{equation}

The communication cost of all nodes is then 
\begin{equation}
\begin{aligned}
N(1 + (C-1)\alpha \frac{N}{CK} ((1 -\mu) p + \mu C))d. &= ( ((1 -\mu) p + \mu C) \frac{\alpha N(C-1)}{CK}  + 1)Nd.\\
&= ( ((1 -\mu) p + \mu C) \frac{\alpha N(C-1)}{CK}  + 1)Nd.\\
&= (  \frac{ (1 -\mu)\alpha N(C-1)}{CK} p  +  \frac{\mu \alpha N(C-1)}{C} + 1)Nd.
\end{aligned}
\end{equation}
\subsection{Server sends to clients}

Since the aggregations of neighbor features have been calculated in the server, it then needs to send the aggregations back to clients. 

For $1$-hop communication, each client requires the aggregations of neighbors ($1$-hop) of its local nodes. The communication cost equals to the number of local nodes times the size of the node feature is given by
\begin{equation}
    \sum_{k=1}^{K} |\mathcal{V}_k| d = Nd.
\end{equation}

For $2$-hop communication, each client requires the aggregations of $2$-hop neighbors of its local nodes, in which the cost equals to the number of $1$-hop neighbors times the size of the node feature,
\begin{equation}
    \sum_{k=1}^{K} |\mathcal{N}_{\mathcal{V}_k}| d
\end{equation}

The number of neighbors in partial i.i.d setting for client $k$ account for the number of local nodes, 1-hop, 2-hop neighbors and taking into account of parameters $\alpha$ and $\mu$ is 
\begin{equation}
\begin{aligned}
\frac{N}{C} + \frac{C-1}{C} N (1 - (1-\alpha)^{\frac{Np}{CK}}(1-\mu \alpha)^{\frac{N(C-p)}{CK}}) & \approx \frac{N}{C} +  \frac{C-1}{C} N (\alpha \frac{N}{CK} ( (1 -\mu) p + \mu C))\\
&= \frac{N}{C} +  \frac{C-1}{C} N (\alpha \frac{N}{CK}  (1 -\mu) p + \alpha \frac{N}{C}  \mu)\\
\end{aligned}
\end{equation}

Then the number of neighbors in partial i.i.id for all clients is given by the sum over all clients
\begin{equation}
\begin{aligned}
{N} + (C-1) N (1 - (1-\alpha)^{\frac{Np}{CK}}(1-\mu \alpha)^{\frac{N(C-p)}{CK}}) & \approx {N} +  (C-1) N  (\alpha \frac{N}{CK}  (1 -\mu) p + \alpha \frac{N}{C}  \mu )\\
\end{aligned}
\end{equation}

The communication cost considering all local nodes and neighbors is then
\begin{equation}
\begin{aligned}
({N} +  (C-1) N  (\alpha \frac{N}{CK}  (1 -\mu) p + \alpha \frac{N}{CK}  \mu C))d &= (1 + (C-1)(\alpha \frac{N}{CK}  (1 -\mu) p + \alpha \frac{N}{C}  \mu ))Nd\\
= (1 + (C-1)\alpha \frac{N}{CK}  (1 -\mu) p + \mu \alpha (C-1) \frac{N}{C}  )Nd\\
\end{aligned}
\end{equation}

For $L$-hop communication, each client requires the aggregations of $L$-hop neighbors of its local nodes, which equals to the number of $(L-1)$-hop neighbors times the size of the node feature,
\begin{equation}
    \sum_{k=1}^{K} |\mathcal{N}_{\mathcal{V}_k}^{L-1}| d.
\end{equation}

\section{Negative Social Impacts of the Work}

We believe that our work overall may have a \emph{positive} social impact, as it helps to protect user privacy during federated training of GCNs for node-level prediction problems. However, by enabling such training to occur without compromising privacy, there is a chance that we could enable improved training of models with negative social impact. For example, models might more accurately classify users in social networks due to their ability to leverage a larger, cross-client dataset of users in the training. Depending on the model being trained, these results could be used against such users, e.g., targeting dissidents under an authoritarian regime. We believe that such negative impacts are no more likely than positive impacts from improved training, e.g., allowing an advertising company to send better products to users through improved predictions of what they will like. This work itself is agnostic to the specific machine learning model being trained.